\documentclass[10pt,twocolumn,letterpaper]{article}

\usepackage{wacv}
\usepackage{times}
\usepackage{epsfig}
\usepackage{graphicx}
\usepackage{amsmath}
\usepackage{amssymb}

\usepackage{algpseudocode}
\usepackage{algorithm}
\usepackage{amsthm}
\usepackage{color}
\usepackage[normalem]{ulem}
\usepackage{marginnote}

\usepackage[firstpage=true,color=black,placement=top,scale=1,opacity=1,vshift=-1cm]{background}

\SetBgContents{
\begin{minipage}{2\linewidth}
\small
\copyright 2016 IEEE. Personal use of this material is permitted. Permission from IEEE must be obtained for all other uses, in any current or future media, including reprinting/republishing this material for advertising or promotional purposes, creating new collective works, for resale or redistribution to servers or lists, or reuse of any copyrighted component of this work in other works.
\\
A definitive version was published in \textbf{2016 IEEE Winter Conference on Applications of Computer Vision (WACV)} and is available at \url{https://dx.doi.org/10.1109/WACV.2016.7477729}
\end{minipage}
}

\DeclareMathOperator*{\divergence}{div}

\usepackage[pagebackref=true,breaklinks=true,letterpaper=true,colorlinks,bookmarks=false]{hyperref}

\wacvfinalcopy % *** Uncomment this line for the final submission

\def\wacvPaperID{451} % *** Enter the wacv Paper ID here

\ifwacvfinal\fi
\begin{document}

%%%%%%%%% TITLE
\title{Variational Multi-Phase Segmentation using High-Dimensional Local Features}

% Authors at the same institution
\author{Niklas Mevenkamp \hspace{2cm} Benjamin Berkels \\
RWTH Aachen University\\
{\tt\small \{mevenkamp,berkels\}@aices.rwth-aachen.de}
}
% Authors at different institutions
%\author{First Author \\
%\and
%Second Author \\

\maketitle
\ifwacvfinal\thispagestyle{empty}\fi

%%%%%%%%% ABSTRACT
\begin{abstract}
   We propose a novel method for multi-phase segmentation of images based on high-dimensional local feature vectors. While the method was developed for the segmentation of extremely noisy crystal images based on localized Fourier transforms, the resulting framework is not tied to specific feature descriptors.
   For instance, using local spectral histograms as features, it allows for robust texture segmentation. The segmentation itself is based on the multi-phase Mumford-Shah model. Initializing the high-dimensional mean features directly is computationally too demanding and ill-posed in practice. This is resolved by projecting the features onto a low-dimensional space using principle component analysis. The resulting objective functional is minimized using a convexification and the Chambolle-Pock algorithm. Numerical results are presented, illustrating that the algorithm is very competitive in texture segmentation with state-of-the-art performance on the Prague benchmark and provides new possibilities in crystal segmentation, being robust to extreme noise and requiring no prior knowledge of the crystal structure.
\end{abstract}

%%%%%%%%% BODY TEXT
\section{Introduction}
Image segmentation, i.e. the task of decomposing an image into disjoint regions that are roughly homogeneous in a suitable sense, is one of the fundamental image processing problems. If three or more regions are sought, one speaks of multi-phase segmentation.
This problem has been studied thoroughly in the literature and entirely different concepts have been put forward as the basis for image segmentation, such as fuzzy region competition \cite{LiNgZe10}, contour detection \cite{ArMaFo11}, random walks \cite{CoXuGr12}, markov random fields \cite{LiBiPl12}, just to name a few. Due to the variety of proposed methods, providing a comprehensive list is beyond the scope of this article, but we refer the interested reader to \cite{VaSa12}. Then there is, of course, the class of variational approaches based on the famous Mumford-Shah energy \cite{MuSh89}.

The most straight-forward application in multi-phase segmentation is to divide images into regions based on their gray or color intensities \cite{ChJiSu01}. A more complex task is to segment images based on their local \textit{structure}. This has applications in texture segmentation \cite{ReDu93}, as well as many medical applications, such as the segmentation of blood vessels \cite{FrReHo12}. Algorithms for structure classification and segmentation usually extract local features from the image, which analyze important properties of the structures of interest, such as the image intensity, position and orientation of edges, or the local frequency spectrum \cite{TaZhSh13}. In the case of texture segmentation, Gabor filters are arguably the most popular source of feature discrimination \cite{WeHiDu96}, often combined with other filters in so-called \textit{local spectral histograms} \cite{LiWa06}. Other methods rely on linear transforms, such as the short-time Fourier transform \cite{BaGi96}, wavelet transforms \cite{ChKa02}, or, more recently, the Stockwell transform \cite{DrStMi09}. While the part of this paper on texture segmentation uses well-proven spectral histograms to recognize regions, it differs from established methods by their integration into a variational framework, allowing to control the regions' connectedness.

Dealing with complex structures, such as textures, often implies high-dimensionality of the parameters describing the problem. However, in image segmentation, one is mostly interested in classifying structure into a few categories, potentially allowing for lower-dimensional representations. Dimension reduction of high-dimensional data is an immensely broad topic \cite{MaPoHe09} and finds applications in many different areas of research. There exist different approaches, but the most widely used techniques are arguably clustering \cite{PaHaLi04} and principal component analysis (PCA) \cite{KaLe97}. The latter two are connected in the sense that the relaxed solution of $k$-means, one of the most popular clustering algorithms, is given by principal components \cite{DiHe04}. PCA has been investigated in the context of variational image segmentation before, both as a means for dimension reduction \cite{PaFe03} and to increase the contrast of color-texture indicators in natural images \cite{HaFeBa13}.

In materials science, an important application of structure-based segmentation is the analysis of crystals. Available methods are based on variational minimization of Mumford-Shah energies that require the local stencil of a reference crystal as prior knowledge \cite{BeRaRu07,BoBeRu10,ElWi14a}.

\subsection{Key contributions}
\begin{itemize}
	\item a widely applicable framework for image segmentation by structure is discussed, including a novel combination of PCA of high-dimensional features, Mumford-Shah and a robust initialization strategy, which allows for a broad choice of feature descriptors
	\item the framework is shown to work very well, even for extremely noisy data, in crystal segmentation, where it generalizes existing methods in the sense that no a-priori information about the crystals is required
\end{itemize}

%------------------------------------------------------------------------
\section{Variational multi-phase segmentation}
\label{sec:VarSeg}
In this section, we briefly recall the Mumford-Shah model \cite{MuSh89} for multi-phase segmentation based on suitable indicator functions. Furthermore, we recall a convexification approach that enables an efficient numerical minimization of the model. Let $\Omega = [0,1]^2$. The task is to divide $\Omega$ into pairwise disjoint regions $\Omega_l$, $l=1,\dots,k$ based on given indicator functions $f_1, \dots, f_k : \Omega \rightarrow \mathbb{R}_{\geq 0}$. $f_l(x)$ can be interpreted as the cost of putting a point $x\in\Omega$ into the set $\Omega_l$. For instance, if an image $g:\Omega\rightarrow\mathbb{R}$ is supposed to be segmented based on its gray values, possible indicator functions are $f_l(x)=(g(x)-c_l)^2$. Here, $c_l$ is the average gray value of $g$ in the $l$-th region.

A segmentation of $\Omega$ based on the indicator functions that guarantees a certain regularity of the segments can be achieved by minimizing the Mumford-Shah energy \cite{MuSh89}:
\begin{equation}
	\label{eq:ms}
	\min_{(\Omega_l)_{l=1}^k} \sum_{l=1}^k \left\lbrace \int_{\Omega_l} f_l\,\mathrm{d}x + \lambda \text{Per}(\Omega_l,\Omega) \right\rbrace.
\end{equation}
Here, $\text{Per}(\Omega_l,\Omega)$ denotes the perimeter of the set $\Omega_l$ in $\Omega$~\cite{AmFuPa00}. Roughly speaking, the perimeter is the length of the boundary of $\Omega_l$, not counting the parts of the boundary that are also on the boundary of $\Omega$.
This problem is hard to address numerically since the unknown variables are sets. In particular, its discrete counterpart, known as Pott's model, is NP-hard. Thus, various convex relaxation approaches have been proposed in the past. For the sake of simplicity, we use one of the most straightforward approaches, given in \cite{ChristopherZach08}. Let us stress that our framework does not rely on this particular choice, but can also be combined with more sophisticated convexification approaches. Let
\begin{equation}
	\label{eq:mscr}
	E[u] := \sum_{l=1}^k \left\lbrace \int_{\Omega} f_l u_l \,\mathrm{d}x + \lambda |u_l|_{\text{TV}(\Omega)} \right\rbrace
\end{equation}
where $u\in \mathcal{U}$ is a vector valued labeling function and
\begin{equation}
	\label{eq:admiss}
	\mathcal{U} := \left\lbrace u \in \text{BV}(\Omega)^k : u \geq 0 \wedge \sum_{l=1}^k u_l = 1 \text{ a.e. in } \Omega \right\rbrace
\end{equation}
is the admissible set. Here,
\begin{equation}
	\label{eq:tv}
	|u|_{\text{TV}(\Omega)} := \sup_{p \in C^1_c(\Omega,\mathbb{R}^2) \atop \|p\|_\infty \leq 1} \int_\Omega u \divergence p
\end{equation}
denotes the total variation and $\text{BV}(\Omega)$ is the space of functions of bounded variation, i.e. the space of Lebesgue integrable functions with finite total variation. Then, the convex relaxation of \eqref{eq:ms} is to minimize $E$ over the set $\mathcal{U}$.
The minimizer $u^*$ can be interpreted as a soft segmentation and can be converted into a hard segmentation by setting $\Omega_l := \lbrace x \in \Omega : u^*_l(x) \geq u^*_j(x) \, \forall j \neq l \rbrace$.

In order to address this minimization numerically, a discretization of the energy \eqref{eq:mscr} and the admissible set \eqref{eq:admiss} is required. Let $X = (x_i)_{i=1}^n \in \mathbb{R}^{2 \times n}$ be a regular 2D pixel grid. We use piecewise constant approximations $f_{li} = f_l(x_i)$ and $u_{li} = u_l(x_i)$ for $i=1,\dots,n$. The corresponding column vectors and matrices of all pixel values are denoted by a boldface letter, e.g.\ $\boldsymbol u=(\boldsymbol u_1,\ldots,\boldsymbol u_k)\in\mathbb{R}^{n\times k}$. Furthermore, we denote with $K:\mathbb{R}^n\rightarrow \mathbb{R}^{2\times n}$ the discrete gradient operator corresponding to the grid $X$ and forward differences. Using this operator to discretize the total variation \eqref{eq:tv}, the minimization of the discretized energy \eqref{eq:mscr} can be posed as the following discrete saddle point problem:
\begin{equation}
	\label{eq:mscrd}
	\min_{\boldsymbol u\in\mathcal{U}_h} \max_{\hat p_{li} \atop \|\hat p_{li}\| \leq 1} \sum_{l=1}^k \sum_{i=1}^n\left\lbrace f_{li} u_{li} + \lambda \langle(K \boldsymbol u_l)_i, \hat p_{li}\rangle\right\rbrace,
\end{equation}
where $\mathcal{U}_h$ is the discrete counterpart of $\mathcal{U}$ and $\hat p_{li}=(p_{li1},p_{li2})$ discretizes $p$ from \eqref{eq:tv} for $u_l$ at node $x_i$.
Problems of this form can be solved with the Chambolle-Pock algorithm~\cite{ChPo11}, summarized in Algorithm~\ref{alg:cp}.
\begin{algorithm}[b]
	\caption{Chambolle-Pock Type 1}
	\label{alg:cp}
	\begin{algorithmic}
		\State $\hat p^{(0)}_{lij} = 0, \, l=1,\dots,k, i=1,\dots,n, j=1,2$
	    \State $u^{(0)}_{li} = \frac{1}{k}, \, l=1,\dots,k, i=1,\dots,n$
	  	\State $\hat{\boldsymbol{u}}^{(0)} = \boldsymbol u^{(0)}$
	    \Repeat
	    	\State $\hat p^{(t+1)} = R_1(\hat p^{(t)} + \sigma K \boldsymbol u^{(t)})$
	    	\State $\hat{\boldsymbol{u}}^{(t+1)} = R_2(\hat{\boldsymbol{u}}^{(t)} - \tau K^* \hat p^{(t+1)})$
	    	\State $\boldsymbol u^{(t+1)} = \hat{\boldsymbol{u}}^{(t+1)} + \theta (\hat{\boldsymbol{u}}^{(t+1)} - \hat{\boldsymbol{u}}^{(t)})$
		\State $t \leftarrow t+1$
	    \Until{$\|\boldsymbol u^{(t+1)}-\boldsymbol u^{(t)}\| < \epsilon$ or $t > t_\text{max}$}
	\end{algorithmic}
\end{algorithm}
The required resolvent operators are given by
\begin{align}
    \label{eq:dualRes} R_1(\boldsymbol p) &= \left(p_{lij} / \max\lbrace \| \hat p_{li}\|, 1 \rbrace \right)_{lij},\\
    \label{eq:primRes} R_2(\boldsymbol u) &= \pi_{\mathcal{U}_h}\left(\boldsymbol u-\textstyle{\frac{\tau}{\lambda}} \boldsymbol f\right).
\end{align}
Here, $\boldsymbol f=(\boldsymbol f_1,\ldots,\boldsymbol f_k)$ and $\pi_{\mathcal{U}_h}(\boldsymbol u)$ denotes the orthogonal projection of $\boldsymbol u$ onto the set $\mathcal{U}_h$. This projection can be calculated with $\mathcal{O}(k)$ operations using an iterative algorithm described in \cite{Mi86}. In this work, all numerical experiments use the parameters $\sigma=\tau=\frac{1}{8}, \theta=0.7, \epsilon=0.001, t_\text{max} = 10000$. The regularization parameter is chosen as $\lambda = 0.01$ (Table~\ref{table:texturePragueICPR2014}), $\lambda = 0.005$ (Table~\ref{table:textureOutex}, Figure~\ref{fig:textureOutex}) and $\lambda = 25$ (Figure~\ref{fig:crystals}).

%-------------------------------------------------------------------------
\section{Local features for structure characterization}
\subsection{Description and relation to Mumford-Shah}
Our aim is to provide a method to segment images into regions of different structure based on the information from local features. In the discrete setting, local features corresponding to a pixel $x_i$ are encoded in the values of an input image $\boldsymbol g$ in a $(2s+1) \times (2s+1)$ window $W_s(x_i)$ centered at $x_i$.
Here, $s\in\mathbb{N}$ determines the scale that is still considered to be local. From these values, features are extracted by an operator of the form $\mathcal{F}: \mathbb{R}^{(2s+1)^2} \rightarrow \mathbb{R}^m$, that should fulfill certain properties, which we will detail later. Applying $\mathcal{F}$ to the $(2s+1) \times (2s+1)$ matrix $\boldsymbol g(W_s(x_i))$ containing the image pixel values in the window $W_s(x_i)$ gives the feature vector corresponding to the pixel $x_i$:
\begin{equation}
	\label{eq:featureExtractor}
	\mathcal{F}[\boldsymbol g](x_i) := \mathcal{F}(\boldsymbol g(W_s(x_i))).
\end{equation}
Let $\Omega_l^* \subset X$, $l=1,\dots,k$ denote the sought discrete regions, i.e.\ the true sets of pixels belonging to the different structure regions. Then, a suitable feature extractor (as defined in \eqref{eq:featureExtractor}) for discriminating regions of different structures can be characterized by the following two properties:
\begin{align}
	\label{eq:prop1} \max\limits_{l=1,\dots,k} \max\limits_{x,x^\prime \in \Omega_l^*} \|\mathcal{F}[\boldsymbol g](x) - \mathcal{F}[\boldsymbol g](x^\prime)\| &\text{ is small}, \\
	\label{eq:prop2} \min\limits_{\substack{l,l^\prime=1,\dots,k\\l\neq l^\prime}} \min\limits_{x \in \Omega_l^*, x^\prime \in \Omega_{l^\prime}^*} \|\mathcal{F}[\boldsymbol g](x) - \mathcal{F}[\boldsymbol g](x^\prime)\| &\text{ is large,\!}
\end{align}
i.e.\ local features should vary as little as possible within each region and offer as much contrast as possible between different regions. Examples for robust feature extractors for texture and crystal segmentation will be discussed in Section~\ref{sec:applications}.

Given a suitable feature extractor and the true mean features within the different structure regions
\begin{equation}
	\boldsymbol c = \bigg( \frac{1}{|\Omega_l^*|} \sum_{x \in \Omega_l^*} \mathcal{F}[\boldsymbol g](x) \bigg)_{l=1}^k \in \mathbb{R}^{m \times k},
\end{equation}
the following indicator can be used for segmentation in \eqref{eq:mscrd}:
\begin{equation}
	\label{eq:indicator}
	f_{li} := \|\mathcal{F}[\boldsymbol g](x_i) - \boldsymbol c_l\|^2.
\end{equation}
In practice, the mean values $\boldsymbol c$ are of course unknown. However, given some approximate guess $\boldsymbol c^{\{t\}}$ for the mean values, Algorithm~\ref{alg:cp} can be applied, resulting in a segmentation $\boldsymbol u^{\{t\}}$. Then, the following update rule can be applied to refine the mean features
\begin{equation}
	\label{eq:means}
	\boldsymbol c^{\{t+1\}}_l = \left.{\sum_{i=1}^n \mathcal{F}[\boldsymbol g](x_i) u^{\{t\}}_{li}}\middle/{\sum_{i=1}^n u^{\{t\}}_{li}}\right..
\end{equation}
This way, given some initial guess $\boldsymbol c^{\{0\}}$, both the segmentation and the mean features can be refined in an alternating fashion. Note that we use curly brackets instead of round ones for the index here, to differentiate between the iterations within Algorithm~\ref{alg:cp} and these outer iterations. Unfortunately, the result of this alternating minimization strategy depends heavily on the initial guess $\boldsymbol c^{\{0\}}$. In the literature, it is often suggested to approximate $\boldsymbol c^{\{0\}}$ via clustering, which is equivalent to minimizing \eqref{eq:ms} for $\lambda = 0$ with \eqref{eq:indicator} as indicator and with respect to both the regions $\Omega_l$ and $\boldsymbol c$. This clustering problem is NP-hard itself, but efficient iterative solvers, such as $k$-means, are available and have proven to work well in the case of low-dimensional indicator functions (e.g. in color segmentation) \cite{ChJiSu01}. However, robust feature extractors suitable for structure discrimination tend to be high-dimensional ($m$ greater than $100$ or even $1000$). In this case, clustering becomes unfeasible in practice, because the available solvers are likely to get stuck in undesired local minima when applied in such high dimensions.

\subsection{Dimension reduction and decorrelation}
Clustering of high-dimensional data is a well studied problem in the literature \cite{PaHaLi04}. It has been noted that often many of the dimensions are irrelevant for the core information expressed by a given data set and that they might mask the essential clusters due to noise. Therefore, several approaches for subspace clustering have been proposed to address this problem \cite{KrKrZi09}. In our context, dimension reduction and decorrelation via principal component analysis (PCA) should work well: given a feature extractor $\mathcal{F}$ fulfilling \eqref{eq:prop1} \& \eqref{eq:prop2}, $\left(\mathcal{F}[\boldsymbol g](x_i)\right)_{x \in \Omega_l^*}$ is of low variance for any $l\in\{1,\dots,k\}$ and, compared to this, for $l \neq l^\prime$ the set $\left(\mathcal{F}[\boldsymbol g](x), \mathcal{F}[\boldsymbol g](x^\prime)\right)_{x \in \Omega_l^*, x^\prime \in \Omega_{l^\prime}^*}$ is of high variance.

Performing PCA on the matrix of mean-centralized features $A = (\mathcal{F}[\boldsymbol g](x_1) - \boldsymbol\mu[\boldsymbol g], \dots, \mathcal{F}[\boldsymbol g](x_n) -  \boldsymbol\mu[\boldsymbol g]) \in \mathbb{R}^{m \times n}$ with $\boldsymbol\mu[\boldsymbol g] = \frac{1}{n} \sum_{i=1}^n \mathcal{F}[\boldsymbol g](x_i)$, results in a lower-dimensional coefficient representation $\boldsymbol\alpha^{(r)} = (U^{(r)})^T A \in \mathbb{R}^{r \times n}$, where $U^{(r)} \in \mathbb{R}^{m \times r}$ is the matrix of eigenvectors belonging to the largest $r$ eigenvalues of $AA^T$. Clustering the coefficients $\boldsymbol\alpha^{(r)}$ into $k$ clusters gives a coefficient representation $\boldsymbol\gamma^{(r)} \in \mathbb{R}^{r \times k}$, which results in the initial guess $\boldsymbol c^{\{0\}} = U^{(r)} \boldsymbol\gamma^{(r)}+\boldsymbol\mu[\boldsymbol g]$. Since we need $\boldsymbol c\in \mathbb{R}^{m \times k}$, $r = k$ is a natural choice for dimension reduction.

In \cite{YuWaChe15}, it was noted that the clustering can get stuck in local minima due to effects caused by the inhomogeneity of the features across the boundary between two regions. Unlike purely point-wise indicators ($s=0$), local feature extractors cause points within about half the window size of a region boundary (in 2D space) to spread between the two mean features corresponding to the regions adjacent to the boundary (in coefficient space). In order to prevent the $k-$means minimizer from getting stuck in-between such two clusters, Yuan et al. proposed to disregard such boundary points when clustering by thresholding an edgeness indicator, given by finite differences of the features on the scale of the window size \cite{YuWaChe15}. As this approach is only based on the assumption of homogeneity of the features within each structure region, it can be used for general feature extractors and allows us to adopt this technique for the initial clustering.

While the above resembles a robust method to retrieve an initial value for $\boldsymbol c$ in the full dimensional feature space, the dimension reduction we now have at hand also suggests itself to reduce the noise of the high dimensional feature vectors and to increase their inter-region contrast within the subsequent variational segmentation framework. First of all, let us point out that as $U^{(m)}$ forms an orthonormal basis of $\mathbb{R}^m$, we can express the indicator \eqref{eq:indicator} and thus the fidelity term in \eqref{eq:mscrd} in terms of $\boldsymbol\gamma^{(m)}$ and $\boldsymbol\alpha^{(m)}$:
\begin{equation}
	\label{eq:coeffs}
	\|\mathcal{F}[\boldsymbol g](x_i) - \boldsymbol c_l\|^2 = \|\boldsymbol\alpha^{(m)}_i - \boldsymbol\gamma^{(m)}_l\|^2
\end{equation}
Furthermore, definition \eqref{eq:means} can be rewritten as
\begin{equation}
	\label{eq:meanCoeffs}
	U^{(m)}\boldsymbol\gamma^{(m)}_l = \boldsymbol c_l -\boldsymbol\mu[\boldsymbol g]= U^{(m)} \frac{\sum_{i=1}^n \boldsymbol\alpha^{(m)}_{i} u_{il}}{\sum_{i=1}^n u_{il}}
\end{equation}
i.e. the mean values $\boldsymbol\gamma$ can be updated using the coefficients $\boldsymbol\alpha$ instead of the feature vectors $\mathcal{F}[\boldsymbol g](x_i)$. Reducing the dimension to $r < m$ introduces an error, which can be bounded by the eigenvalues $\lambda_1 \geq \dots \geq \lambda_m$ of $A A^T$:
\begin{equation}
	\label{eq:fidelityError}
	\big| f_{li} - \|\boldsymbol\alpha^{(r)}_i - \boldsymbol\gamma^{(r)}_l\|^2 \big| \leq 2 \textstyle\sum_{j=r+1}^m \lambda_j
\end{equation}
This inequality can be deduced by applying the triangle inequality to the difference of the left- and right-hand side in \eqref{eq:coeffs}, splitting $\boldsymbol{\alpha}^{(m)}_i, \boldsymbol{\gamma}^{(m)}_l$ into $\boldsymbol{\alpha}^{(r)}_i, \boldsymbol{\gamma}^{(r)}_l$ and remaining parts $\bar{\boldsymbol{\alpha}}_i, \bar{\boldsymbol{\gamma}}_l$, as well as representing $\bar{\boldsymbol{\alpha}}_i, \bar{\boldsymbol{\gamma}}_l$ as convex combinations of the columns of $(U^{(m)})^T A$ with non-zero coefficients in columns corresponding to $\lambda_{r+1},\dots,\lambda_m$.

Note that the error in \eqref{eq:fidelityError} can be estimated without calculating all $m$ eigenvalues, which may become computationally expensive when the dimension $m$ becomes large:
\begin{equation}
	\label{eq:fidelityErrorEfficient}
	\textstyle\sum_{j=r+1}^m \lambda_j = \|A\|_F^2 - \textstyle\sum_{j=1}^r \lambda_j.
\end{equation}
Here, $\|A\|_F = \sqrt{\sum_{i=1}^m\sum_{j=1}^n |A_{ij}|^2}$ denotes the Frobenius norm. This way, the error in the fidelity term can still be monitored, when the eigenvectors of $AA^T$ are calculated iteratively, e.g.\ using a deflation type of strategy. We use $r = k$ within the entire framework. Algorithm~\ref{alg:pcamss} summarizes the proposed method. All numerical experiments use $\hat t_\text{max} = 3$. The edgeness threshold on the finite differences before clustering in Algorithm~\ref{alg:pcamss} is chosen as $\delta = 0.5$ (Table~\ref{table:texturePragueICPR2014}), $\delta = 0.25$ (Table~\ref{table:textureOutex}, Figure~\ref{fig:textureOutex}) and $\delta = 1.0$ (Figure~\ref{fig:crystals}).
\begin{algorithm}
	\caption{Variational multi-phase feature segmentation}
	\label{alg:pcamss}
	\begin{algorithmic}
		\State $A = \left(\mathcal{F}[\boldsymbol{g}](x_i)\right)_{i=1}^n$
		\State $\boldsymbol\alpha = (U^{(k)})^T A$
		\State $\Delta_i = \sum_{j : x_j = x_i \pm (0,s) \vee x_j = x_i \pm (s,0)} \|\alpha_i - \alpha_j\|$
		\State $\tilde{\boldsymbol\alpha} = \left( \boldsymbol\alpha_i : \Delta_i < \delta \cdot \frac{1}{n} \sum_{i=1}^n \Delta_i \right)$
		\State $\left(\boldsymbol\gamma_l\right)_{l=1}^k = k$-$\mathrm{means}(\tilde{\boldsymbol\alpha})$
		\State $u^{\{0\}}_{li} = \delta_{l,\arg\min_{l^\prime} \|\boldsymbol\alpha_i - \boldsymbol\gamma_{l^\prime}\|^2}$
 		\State $\boldsymbol{p}^{\{0\}} = 0$
		\Repeat
			\State $f_{li} = \|\boldsymbol\alpha_i - \boldsymbol\gamma_l\|^2, l=1,\dots,k, i=1,\dots,n$
			\State $(\boldsymbol{u}^{\{\hat{t}+1\}}, \boldsymbol{p}^{\{\hat{t}+1\}}) = \mathrm{Algortihm\,1}(\boldsymbol f, \boldsymbol{u}^{\{\hat{t}\}}, \boldsymbol{p}^{\{\hat{t}\}})$
			\State $\boldsymbol\gamma_l = \sum_{i=1}^n \boldsymbol\alpha_i u^{\{\hat{t} + 1\}}_{il} / \sum_{i=1}^n u^{\{\hat{t} + 1\}}_{il}$
		\Until $\hat{t} = \hat{t}_\text{max}$
		\State \Return $\mathrm{mask}_i = \arg\max_{l=1,\dots,k} u^{\{\hat{t}_\text{max} + 1\}}_{il}$
	\end{algorithmic}
\end{algorithm}

\subsection{Properties and advantages of the method}
PCA has been used, for instance, as a concept for dimension reduction of PET data and subsequent variational segmentation \cite{PaFe03}, as well as a tool for increasing the contrast in the region descriptors for natural color-texture images in a variational segmentation framework \cite{HaFeBa13}. Moreover, Yuan et al.~\cite{YuWaChe15} utilized the related concept of singular value decomposition to compute a low-rank factorization of a local spectral histogram based feature matrix and estimate subsequent template features via clustering. We want to stress that the initialization step in the proposed method shares the idea of dimension reduction and clustering of features, albeit differing slightly in the details, and is, in this regard, similar to \cite{YuWaChe15}. However, our work embeds these ideas into a variational segmentation framework, which grants the following two main advantages:

First, the proposed method can be applied to a very general class of feature extractors, since it only relies on the natural properties \eqref{eq:prop1}, \eqref{eq:prop2}. In particular, in contrast to \cite{YuWaChe15}, it does not rely on the assumption that the feature vectors are linear combinations of the mean features in each region (this assumption and its consequences are discussed later in this section). Among others, the generality of our framework allows the usage of globally coupling, convolution based linear transforms. Functions of this type, such as the short-time Fourier transform \cite{BaGi96}, the Stockwell-Transform \cite{DrStMi09}, or different types of wavelet transforms \cite{ChKa02}, have been studied for texture segmentation and shown their performance.

Second, the dimension reduction of the fidelity term helps to increase the degree to which it fulfills \eqref{eq:prop1}, \eqref{eq:prop2}. In particular, incorporating the PCA not only in the initial clustering, but also throughout the entire variational minimization, helps to suppress noise in the fidelity term. In Section~\ref{sec:UnsupCrystSegm}, we will demonstrate how effective this strategy performs in the case of extremely noisy crystal images, using the Fourier transform as the feature extractor.

Unlike \cite{LiWa06, YuWaChe15}, the general applicability of the proposed framework is tied to the need for a regularization of the segment boundaries, which is covered by the Mumford-Shah model. This need arises from an unexpected behavior of the indicator functions near segment boundaries.
Due to the window size, the feature extractor sees a mixture of different segments near boundaries. For general feature extractors, this means that feature vectors near boundaries are not necessarily a linear combination of the cluster centers corresponding to the adjacent segments.
In case $k > 2$, it might happen that the feature vector at a boundary between two regions $\Omega_1^*, \Omega_2^*$ is nearer to the mean feature vector of a third region $\Omega_3^*$ than it is to $\Omega_1^*$ or $\Omega_2^*$ itself.
This means that the indicator $\boldsymbol f$ cannot necessarily identify the correct segment within a distance of $s$ to the sought segments. Note that this effect does not arise for $k=2$. As mentioned above, the perimeter regularization within the Mumford-Shah model addresses this problem for practical purposes. For input data where the regularization alone is not sufficient, we suggest to combine feature extractors of different window sizes.

Beyond this, the proposed method is an extension of \cite{LiWa06, YuWaChe15} in the sense that the decoupling of the coefficient representation from the segmentation allows for an outer iteration to refine the mean features, whereas in \cite{YuWaChe15} the clusters are solely computed from the feature matrix.

Let us point out that, since the method is based on local windows, it has the common limitations inherent to such methods. The feature scale is tied to the window size $s$, so the method can only reliably detect regions that are at least somewhat larger than the window $W_s(x)$. Furthermore, special care has to be taken close to the boundary, where the window $W_s(x)$ leaves the support of the image. Please note that the proposed method enforces the region boundaries to approach the image boundary orthogonally, which is due to the natural boundary conditions in the Euler-Lagrange equation of \eqref{eq:mscr}. This effect can be reduced by introducing ghost cells at the image boundary with a zero extension of all indicators, but it is still noticeable (cf. Figure~\ref{fig:textureOutex}).

%-------------------------------------------------------------------------
\section{Applications and numerical results}
\label{sec:applications}
%-----------------------------------------------------------
\subsection{Texture segmentation}
\label{sec:TextSegm}
Apart from plain gray value or color intensities, among the most thoroughly studied types of structures in image segmentation are textures \cite{IlWh11}. In the image processing sense, a texture essentially consists in a more or less strictly repetitive pattern of the spatial arrangement of the gray or color values in an image. Thus, indicators for texture segmentation need to take into account image information from a whole neighborhood, at least on the scale of the spatial distance between repetitions.
There are two main classes of operators that have been proposed in the literature, namely 1) local spectral histograms combined with a suitable bank of filters and 2) localized linear transforms, both of which fall into the class of feature extractors described earlier. In the context of texture segmentation, we limit our analysis to the first class, while the second class will be utilized for crystal segmentation in the next section.

Local spectral histograms are defined as follows: first a bank of $p$ filters is selected and applied to the image, resulting in a sequence of filtered images $\boldsymbol g_1, \dots,\boldsymbol  g_p$. Then, the feature extractor is defined by
\begin{equation}
	\label{eq:lsh}
	\left(\mathcal{F}_\text{SH}[\boldsymbol g](x)\right)_{ij} = \frac{\sum\limits_{x_k \in W_s(x)}\int_{z_{ij}}^{z_{i,j+1}} \delta(z - g_{ik}))\,\mathrm{d}z}{|W_s(x)|}
\end{equation}
Here, $z_{ij}$, $i=1,\dots,p$, $j=1,\dots,q+1$ define the binning of the histograms and are often chosen such that $z_{i,1} = \min_x g_i(x), z_{i,q} = \max_x g_i(x)$ and equidistant in between.
Thus, the dimension of the extracted feature at every pixel $x$ is $m = pq$.
The most popular filter used in this context is arguably the Gabor filter. Other commonly used filters are Gaussian filters, Laplacian of Gaussian filters, or just the intensity filter (i.e. the identity). For a thorough description of spectral histograms of filtered images and their application to texture segmentation, we refer to \cite{LiWa06}.

In the following, we compare our approach to the \textrm{Outex\_US\_00000} test suite of the \textit{Outex texture database} (\url{http://www.outex.oulu.fi}) and the Prague ICPR2014 contest~\cite{HaMi14} (\url{http://mosaic.utia.cas.cz/icpr2014/}).

%----------------------------------- Outex -----------------------------------
\begin{table*}[t]
    \begin{center}
        \begin{tabular}{lcccccc}
        \hline
        Method & CS $\uparrow$ & O $\downarrow$ & C $\downarrow$ & CA $\uparrow$ & CO $\uparrow$ & CC $\uparrow$ \\
        \hline
        FSEG$^*$-$\mathrm{TxtMerge}$ & \textbf{85.20} $\pm$ 22.7 & *4.76 $\pm$ 3.9 & 6.05 $\pm$ 5.5 & *82.49 $\pm$ 13.4 & *89.23 $\pm$ 9.6 & *88.20 $\pm$ 11.8 \\
        Algorithm~\ref{alg:pcamss} & *80.60 $\pm$ 28.3 & 5.82 $\pm$ 8.3 & \textbf{5.62} $\pm$ 6.8 & \textbf{82.73} $\pm$ 15.5 & \textbf{89.24} $\pm$ 11.3 & \textbf{88.84} $\pm$ 13.3 \\
        FSEG$^*$ & 68.60 $\pm$ 24.3 & \textbf{3.61} $\pm$ 3.2 & *6.04 $\pm$ 8.8 & 70.67 $\pm$ 17.8 & 80.40 $\pm$ 13.1 & 73.51 $\pm$ 18.6 \\
        Clustering & 60.00 $\pm$ 26.8 & 13.36 $\pm$ 8.1 & 15.52 $\pm$ 12.0 & 66.41 $\pm$ 12.5 & 77.54 $\pm$ 10.0 & 77.51 $\pm$ 11.2 \\
        FSEG~\cite{YuWaChe15} & 45.80 $\pm$ 26.0 & 19.29 $\pm$ 34.2 & 17.55 $\pm$ 19.1 & 50.65 $\pm$ 20.6 & 64.95 $\pm$ 16.3 & 52.88 $\pm$ 21.9 \\
        \hline
        \end{tabular}
    \end{center}
    \caption{Gray-scale texture segmentation comparison on the \textrm{Outex\_US\_00000} test suite with known number of segments. FSEG~\cite{YuWaChe15} uses the ICPR2014 code of FSEG with fixed number of segments ($\mathrm{segn = 5}$). FSEG$^*$ and Algorithm~\ref{alg:pcamss} both combine spectral histograms with $11$ bins, window sizes $s=15$ and $s=30$ (stacked with weights 0.8 and 0.2 respectively), and Gabor filters of kernel sizes $\sigma=5,7,9$ and orientations $\theta=0,\frac{1}{2}\pi,\frac{1}{4}\pi,-\frac{1}{4}\pi$. FSEG$^*$-$\mathrm{TxtMerge}$ is the same as FSEG$^*$ but runs without FSEG's $\mathrm{TxtMerge}$ post-processing. Bold face highlights the best, a star the second-best result in each column.}
    \label{table:textureOutex}
\end{table*}
\begin{figure*}[t]
\begin{center}
  \includegraphics[width=0.1325\textwidth]{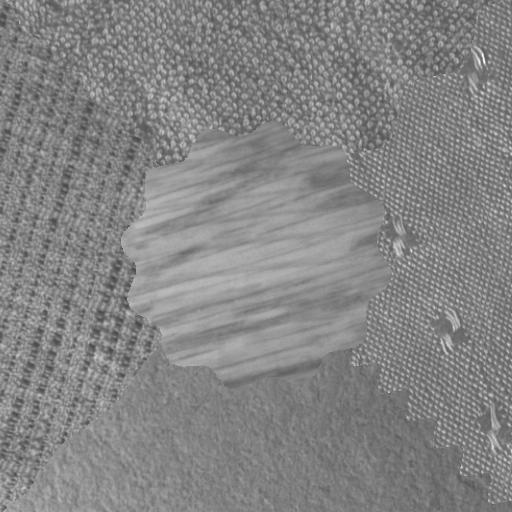}
  \includegraphics[width=0.1325\textwidth]{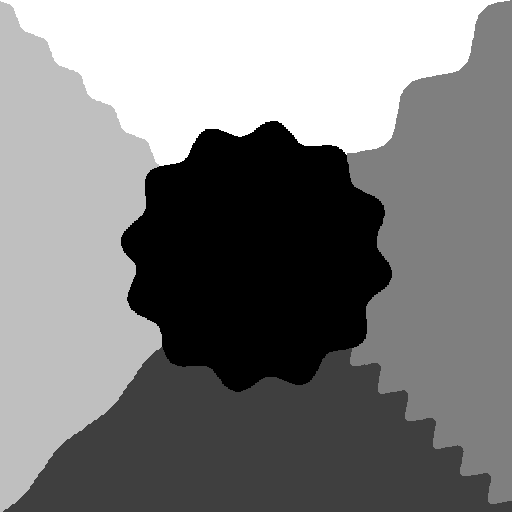}  
  \includegraphics[width=0.1325\textwidth]{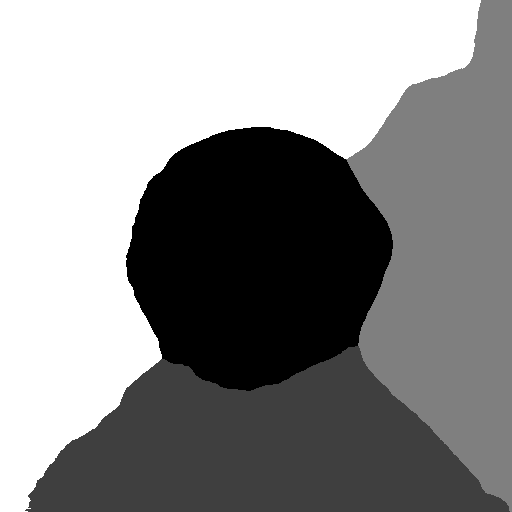}
  \includegraphics[width=0.1325\textwidth]{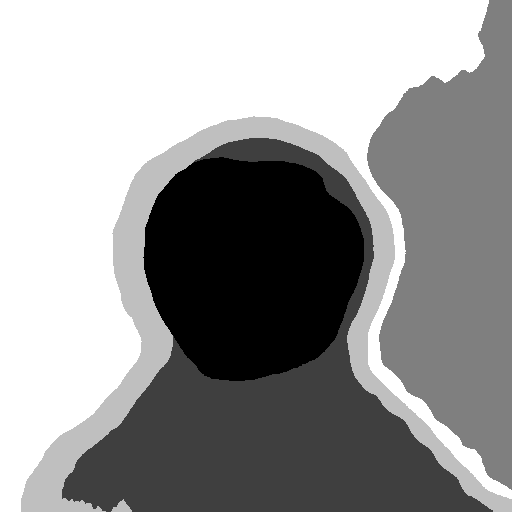}
  \includegraphics[width=0.1325\textwidth]{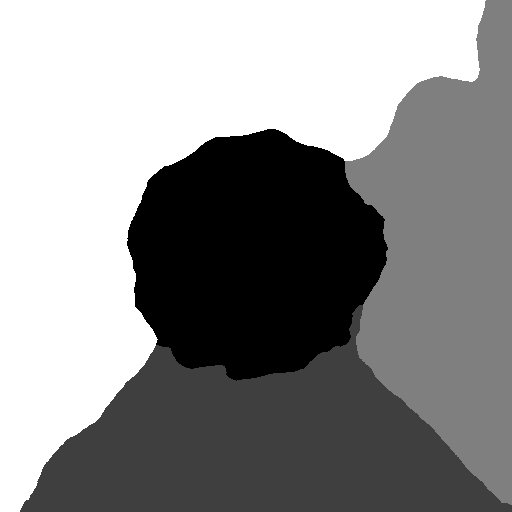}
  \includegraphics[width=0.1325\textwidth]{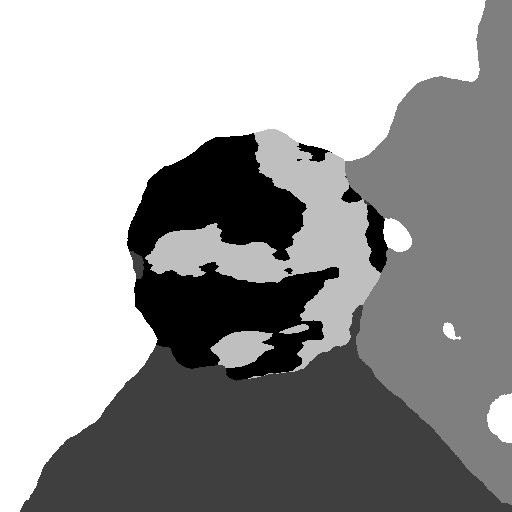}  
  \includegraphics[width=0.1325\textwidth]{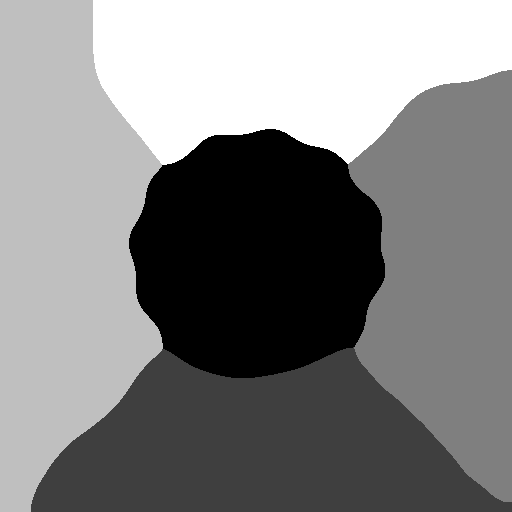}
  \\ \vspace{0.1em}
  \includegraphics[width=0.1325\textwidth]{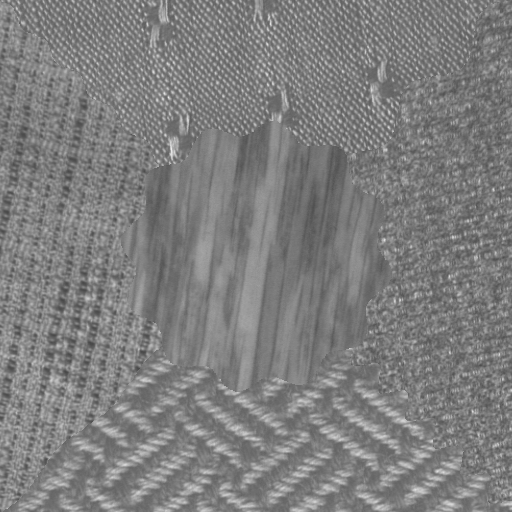}
  \includegraphics[width=0.1325\textwidth]{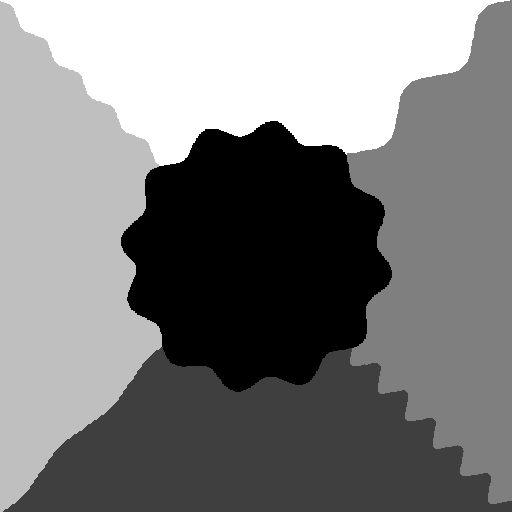}  
  \includegraphics[width=0.1325\textwidth]{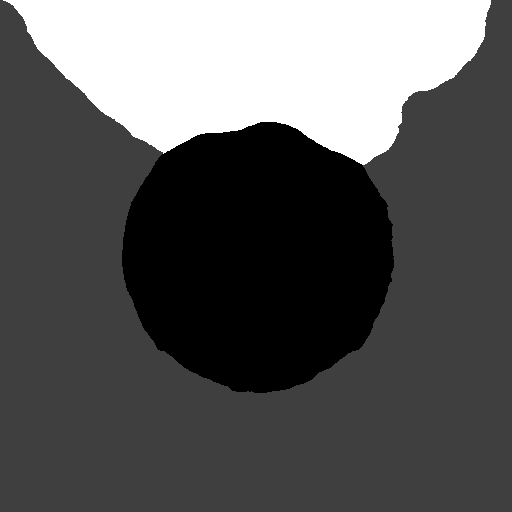}
  \includegraphics[width=0.1325\textwidth]{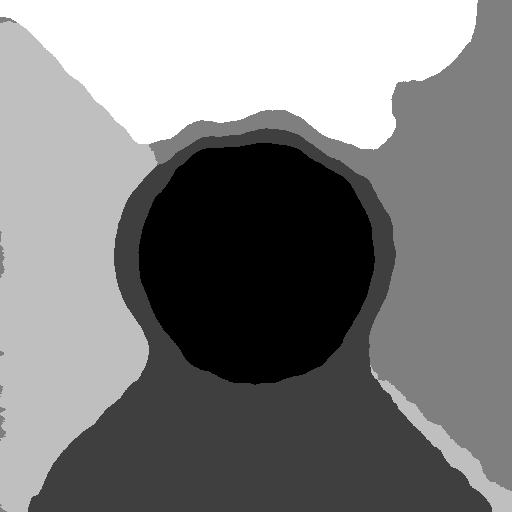}
  \includegraphics[width=0.1325\textwidth]{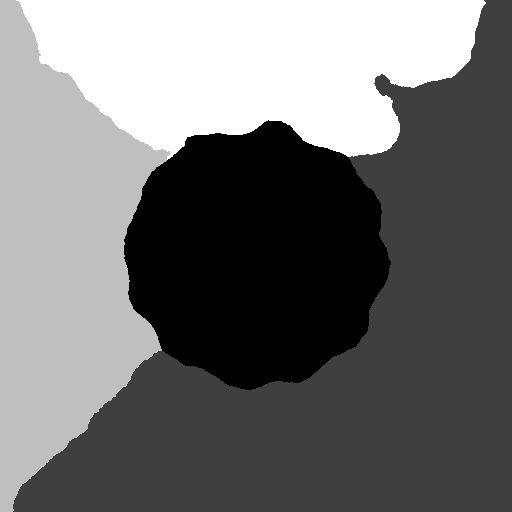}
  \includegraphics[width=0.1325\textwidth]{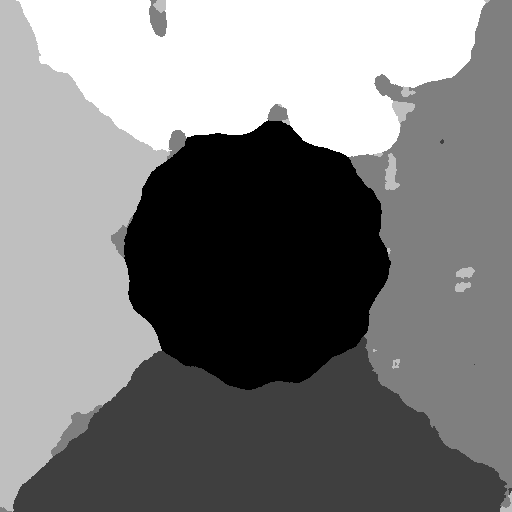}  
  \includegraphics[width=0.1325\textwidth]{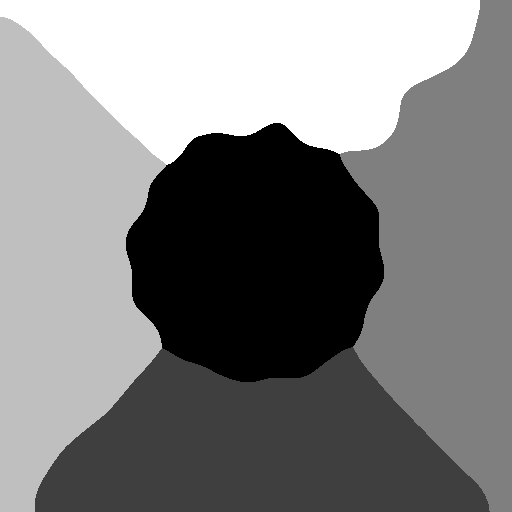}
  \\ \vspace{0.1em}
  \includegraphics[width=0.1325\textwidth]{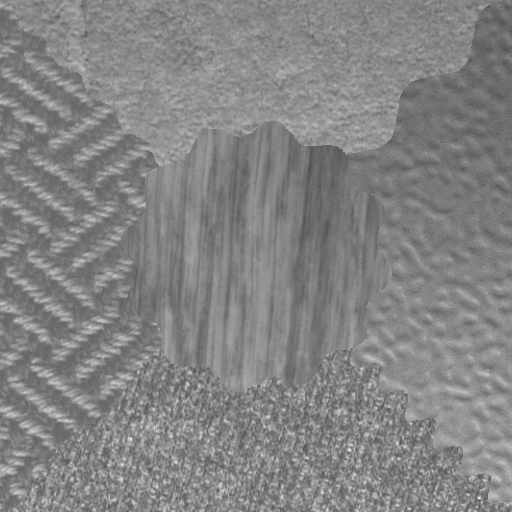}
  \includegraphics[width=0.1325\textwidth]{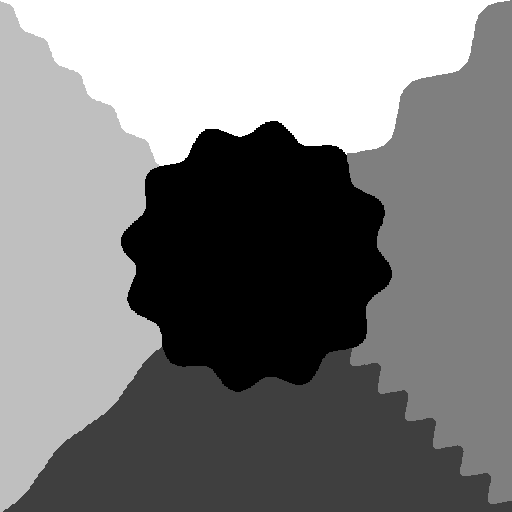}  
  \includegraphics[width=0.1325\textwidth]{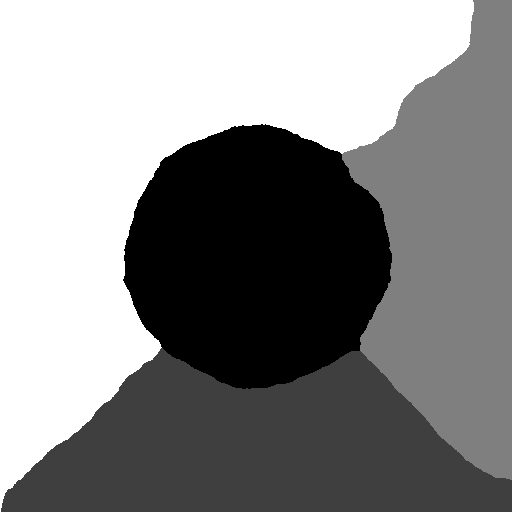}
  \includegraphics[width=0.1325\textwidth]{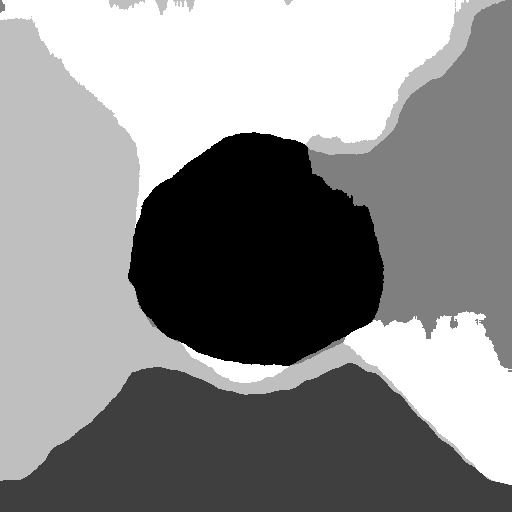}
  \includegraphics[width=0.1325\textwidth]{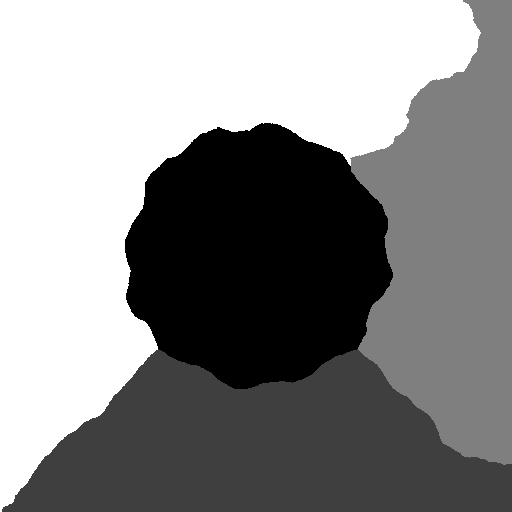}
  \includegraphics[width=0.1325\textwidth]{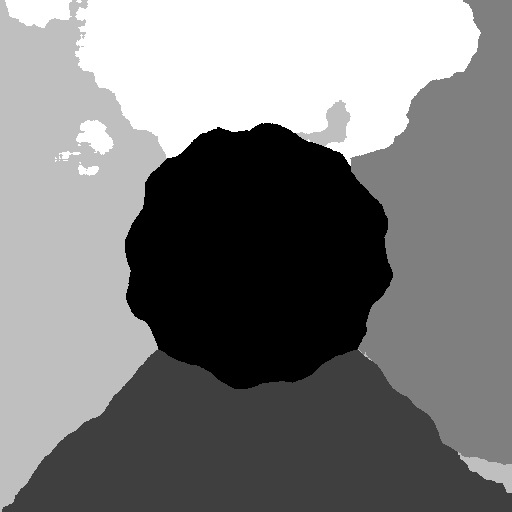}  
  \includegraphics[width=0.1325\textwidth]{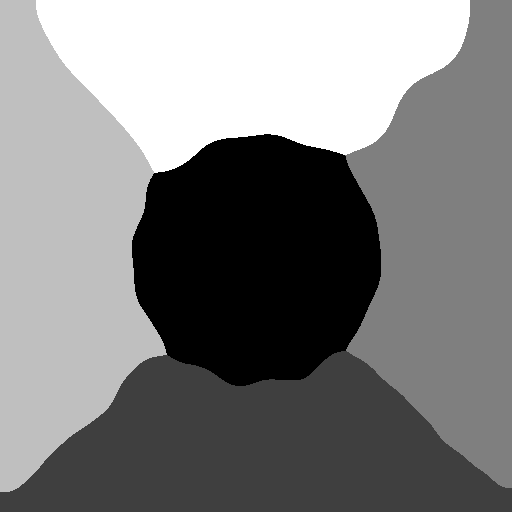}  
\end{center}
   \caption{Segmentations of the first three mosaics from the \textrm{Outex\_US\_00000} test suite. The first column shows the original image, the second the ground truth and the remaining columns the results by FSEG \cite{YuWaChe15}, Clustering, FSEG$^*$, FSEG$^*$-$\mathrm{TxtMerge}$ and Algorithm~\ref{alg:pcamss}.}
\label{fig:textureOutex}
\end{figure*}

On the Outex database, we provide a thorough comparison between the proposed method and \textit{Factorization-Based Texture Segmentation} (FSEG)~\cite{YuWaChe15}. We chose to focus on FSEG here, because it 1) ranks best in the ICPR2014 contest among methods with available code and 2) is similar to our framework.
Table~\ref{table:textureOutex} quantifies the mean segmentation performance and its standard deviation over all 100 texture mosaics from the \textrm{Outex\_US\_00000} test suite. 
Three different versions of FSEG, described in the caption of Table~\ref{table:textureOutex}, are used for this comparison. 
Note that running FSEG without its $\mathrm{TxtMerge}$ post-processing (first row) makes sense in this case, since the number of segments is known. However, this also disables filling holes in the segments, as seen in the sixth column of Figure~\ref{fig:textureOutex}. While FSEG$^*$-$\mathrm{TxtMerge}$ performs best in correct segmentation (CS), and FSEG$^*$ achieves the smallest omission error (O), our method ranks highest in all remaining measures. Note that in Table~\ref{table:textureOutex} we also compare our method to plain clustering in order to evaluate the benefit of 1) the improved initialization strategy via PCA and 2) the subsequent variational optimization including region boundary smoothing. Indeed, the proposed method performs significantly better than plain clustering. Finally, a visual inspection of Figure~\ref{fig:textureOutex} indicates that the proposed method provides a good compromise between fidelity of region boundaries and reduction of artifacts (holes, missing regions).
%----------------------------------- Prague ICPR2014 -----------------------------------
\begin{table*}[t]
    \begin{center}
        \begin{tabular}{lcccccccccc}
        \hline
        Method & CS $\uparrow$ & OS $\downarrow$ & US $\downarrow$ & ME $\downarrow$ & NE $\downarrow$ & O $\downarrow$ & C $\downarrow$ & CA $\uparrow$ & CO $\uparrow$ & CC $\uparrow$ \\
        \hline
        VRA-PMCFA$^\text{\textdagger}$ & \textbf{75.32} & \hspace{-0.5em}*11.95 & \hspace{-0.5em}*9.65 & 4.57 & \hspace{-0.5em}*4.63 & \textbf{4.51} & 8.87 & \textbf{83.50} & \textbf{88.16} & 90.73 \\
        Algorithm~\ref{alg:pcamss}$+\mathrm{TxtMerge}$ & \hspace{-0.5em}*72.27 & 18.33 & \textbf{9.41} & \textbf{4.19} & \textbf{3.92} & \hspace{-0.5em}*7.25 & \textbf{6.44} & \hspace{-0.5em}*81.13 & \hspace{-0.5em}*85.96 & \textbf{91.24} \\
        FSEG~\cite{YuWaChe15} & 69.18 & 14.69 & 13.64 & 5.13 & \hspace{-0.5em}*4.63 & 9.25 & 12.55 & 78.22 & 84.44 & 87.38 \\
        SegTexCol$^\text{\textdagger}$ & 61.19 & \textbf{1.92} & 27.02 & 9.33 & 9.05 & 15.17 & 12.12 & 71.69 & 81.16 & 76.34 \\
        MW3AR8~\cite{Haindl2009} & 53.66 & 51.40 & 14.21 & *5.54 & 6.33 & 19.86 & 84.27 & 70.15 & 75.41 & 89.36 \\
        RS$^\text{\textdagger}$ & 46.02 & 13.96 & 30.01 & 12.01 & 11.77 & 35.11 & 29.91 & 58.75 & 68.89 & 69.30 \\
        Deep Brain~\cite{Felzenszwalb2004} & 36.20 & 41.87 & 53.87 & 7.38 & 9.06 & 47.53 & 99.56 & 49.97 & 62.62 & 70.08 \\
        \hline
        \end{tabular}
    \end{center}
    \caption{Color texture segmentation on the Prague ICPR2014 contest dataset with unknown number of segments. Bold face highlights the best, a star the second-best value in each column, and \textdagger{} indicates that no corresponding publication could be found at the time of writing.}
    \label{table:texturePragueICPR2014}
\end{table*}

Next, we compare the proposed method to results from the Prague ICPR2014 contest~\cite{HaMi14}. Here, we use the same feature extractors as above (on the lightness channel), except that the kernel size $\sigma=9$ is omitted. Therefore, we add an intensity filter on all three channels (in L*a*b color space) to each spectral histogram. The number of segments is estimated as $k=\min\{ k^\prime : \frac{1}{n} \sum_{j=k^\prime+1}^m \lambda_j < \omega\}$ with $\omega = 0.05$. Since this estimate is not precise (even for an optimal choice of $\omega$), we additionally employ FSEG's $\mathrm{TxtMerge}$ post-processing. Table~\ref{table:texturePragueICPR2014} quantifies the mean segmentation quality over all 80 colored texture mosaics from the Prague ICPR2014 contest dataset. While our method produces larger over-segmentation (OS) than the other best-ranked methods, indicating stronger over-estimation of the number of segments, it performs best for under-segmentation (US), indicating a good coverage of all ground truth segments, reflecting the good initialization strategy. Moreover, our method performs second-best for correct segmentation (CS), omission error (O), class accuracy (CA) and correct assignment (CO). Most notably, according to all other presented measures and in total half of them, our method performs best among all competitors.

Note that VRA-PMCFA resolves fine boundary features but produces labeling noise, whereas our method smoothes region boundaries in favor of suppressing labeling noise. Thus, it depends on the application which of the two methods is likely to be more suitable.

%-----------------------------------------------------------
\subsection{Unsupervised crystal segmentation}
\label{sec:UnsupCrystSegm}
A fundamental research topic in materials science is the analysis and modeling of crystals. Modern transmission electron microscopes (TEM) allow for imaging at atomic scale, which makes the crystal grid visible. In a perfect setting, the crystal is given by a Bravais lattice
\begin{equation}
	L_{a_1,a_2} = \lbrace n_1 a_1 + n_2 a_2 : n_1,n_2 \in \mathbb{Z} \rbrace,
\end{equation}
where $a_1,a_2 \in \mathbb{R}^2$ are the lattice vectors defining the orientation and spacing of the crystal. However, crystals of interest usually exhibit a more complicated behavior, like discontinuous orientation changes along so-called \textit{grain boundaries}. The fully automatic analysis of grain geometries in TEM images is subject of ongoing research \cite{ElWi14a}.

Available variational approaches for grain segmentation \cite{BeRaRu07,BoBeRu10,ElWi14a} are built on the assumption that all grains can be characterized through transformations of a local stencil $q_1,\dots,q_N \in \mathbb{R}^2$, corresponding to a reference crystal given by all linear combinations of $a_1, a_2$ with coefficients in $\{-1,0,1\}$. Then, the Mumford-Shah model with an indicator function of the following type can be used \cite{BeRaRu07, BoBeRu10}:
\begin{equation}
\label{eq:oldGrainOrientationIndicator}
 f_l(x) := \textstyle\frac{1}{N} \textstyle\sum_{k=1}^{N} d(g(x), g(x + M(\alpha_l)q_k)),
\end{equation}
Here, $d$ denotes a suitable intensity distance function and $M(\alpha_l) \in \mathbb{R}^{2 \times 2}$ is an orthogonal matrix, rotating the stencil by the angle of the the $l$-th grain relative to the reference.

The need for a-priori knowledge of the reference crystal structure inherent to indicators like \eqref{eq:oldGrainOrientationIndicator} is a severe limitation of available methods.
As we will show, this limitation can be overcome by using our proposed framework with the modulus of the 2D-FFT as the local feature extractor:
\begin{equation}
	\label{eq:fft}
	\mathcal{F}_\text{FFT}[\boldsymbol g](x) = (|\text{FFT}[\boldsymbol g(W_s(x))]_{ij}|)_{ij}.
\end{equation}
Let us assume that the window $W_s(x)$ is large enough to cover at least one period of the crystal in either direction at any location. Then, the modulus of the Fourier transform $\mathcal{F}_\text{FFT}[\boldsymbol g](x)$ automatically encodes the local stencil $\left(M(\alpha_l)q_k\right)_{k=1}^N$ within the positions of Bragg reflections.

Assuming that the window size $s$ matches the period of the crystal and the unit cell is a square, i.e.\ the discrete signals $\boldsymbol g(W_s(x))$ are exactly periodic, the translation of the window across the image causes a phase shift in frequency domain. This phase shift is canceled by the absolute value in the modulus, making the feature extractor $\mathcal{F}_\text{FFT}[\boldsymbol g](x)$ translation invariant inside crystal regions with fixed lattice parameters. Though in practice this assumption is not met, artifacts in frequency domain caused by window boundary effects are easily handled by the perimeter regularization, as long as the window size $s$ is chosen reasonably large. Furthermore, these are also reduced through the proposed dimension reduction of the fidelity term \eqref{eq:coeffs}. Note that crystal images are usually far from periodic at the boundary ($s$ pixels in orthogonal direction) and thus $\mathcal{F}_\text{FFT}[\boldsymbol g](x)$ cannot be reasonably defined there. Here, we simply extend the segmentation constantly to cover the boundary region.

\begin{figure*}
\begin{center}
	\includegraphics[width=0.215\textwidth]{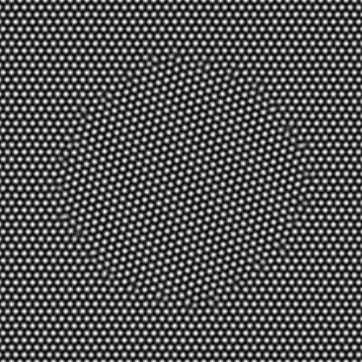}
	\includegraphics[width=0.215\textwidth]{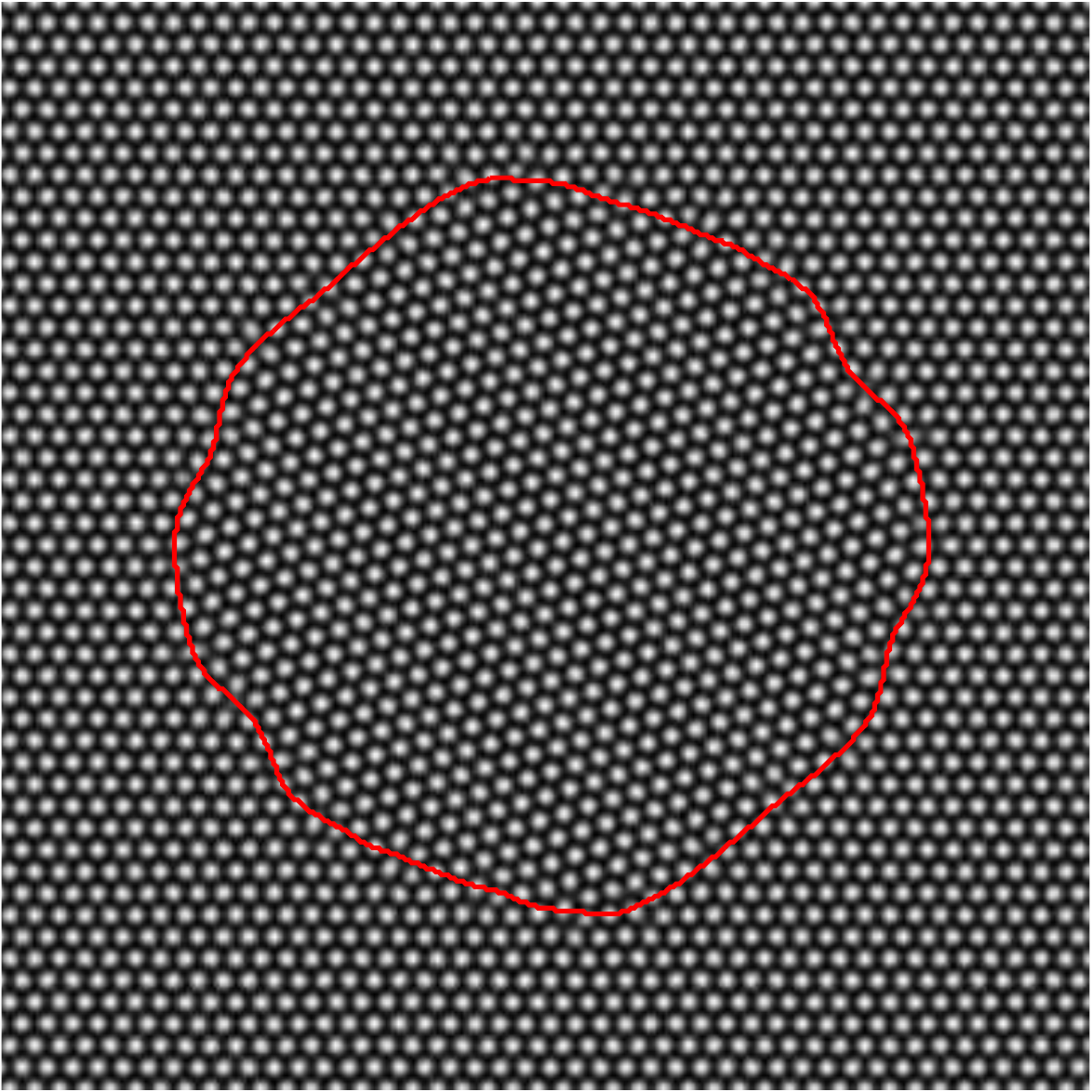}
	\includegraphics[width=0.215\textwidth]{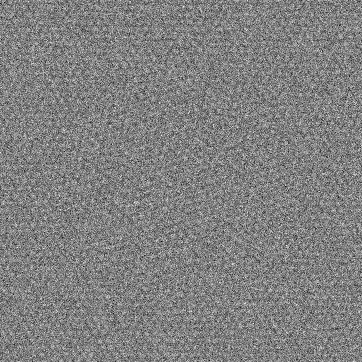}
	\includegraphics[width=0.215\textwidth]{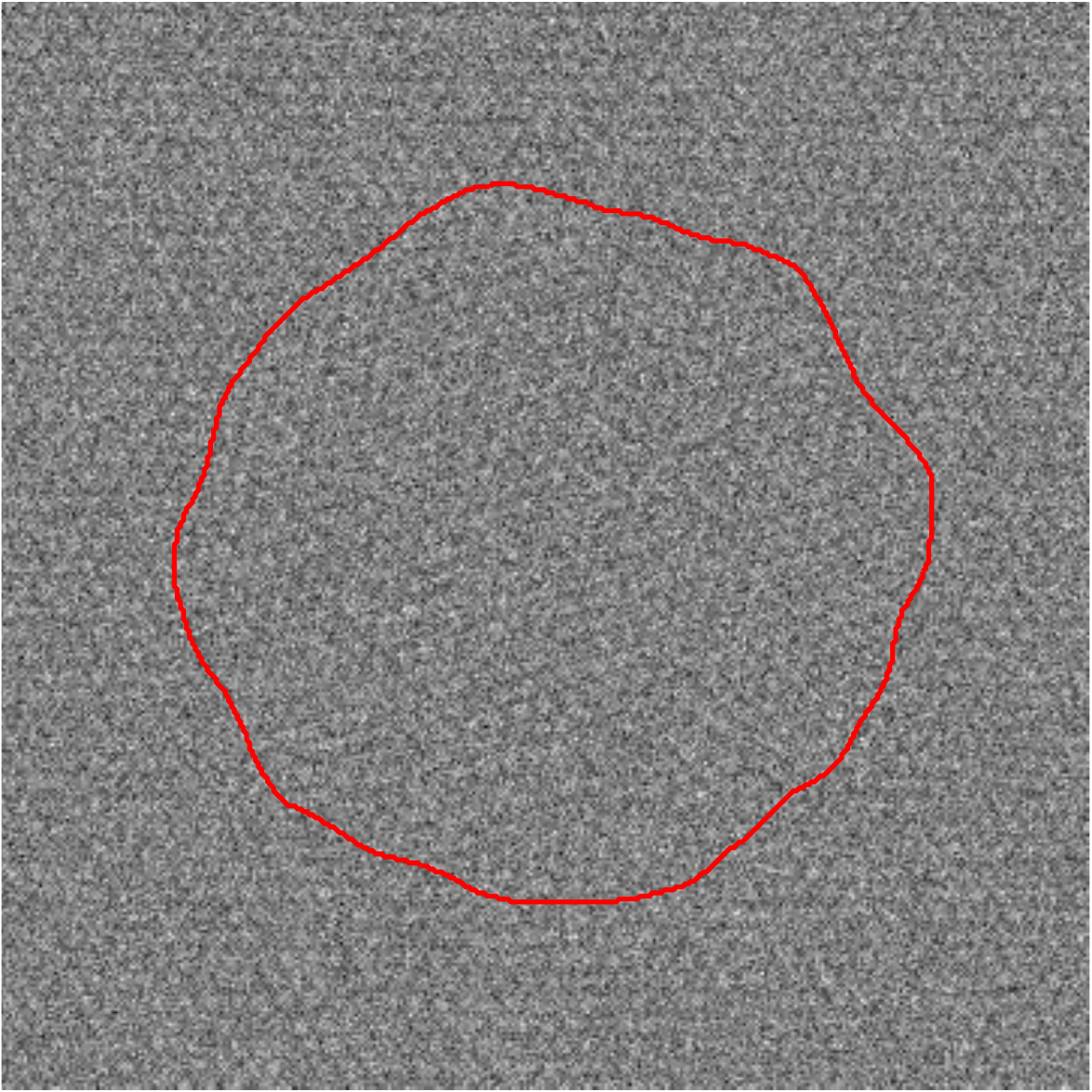}
 	\\ \vspace{0.1em} 
 	\includegraphics[width=0.215\textwidth]{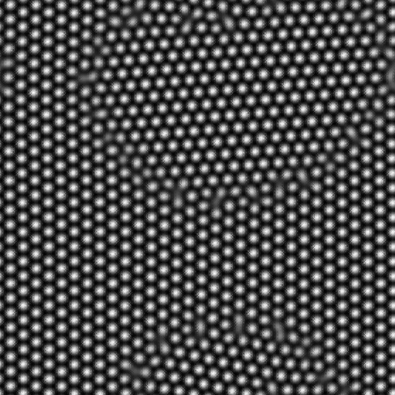}
	\includegraphics[width=0.215\textwidth]{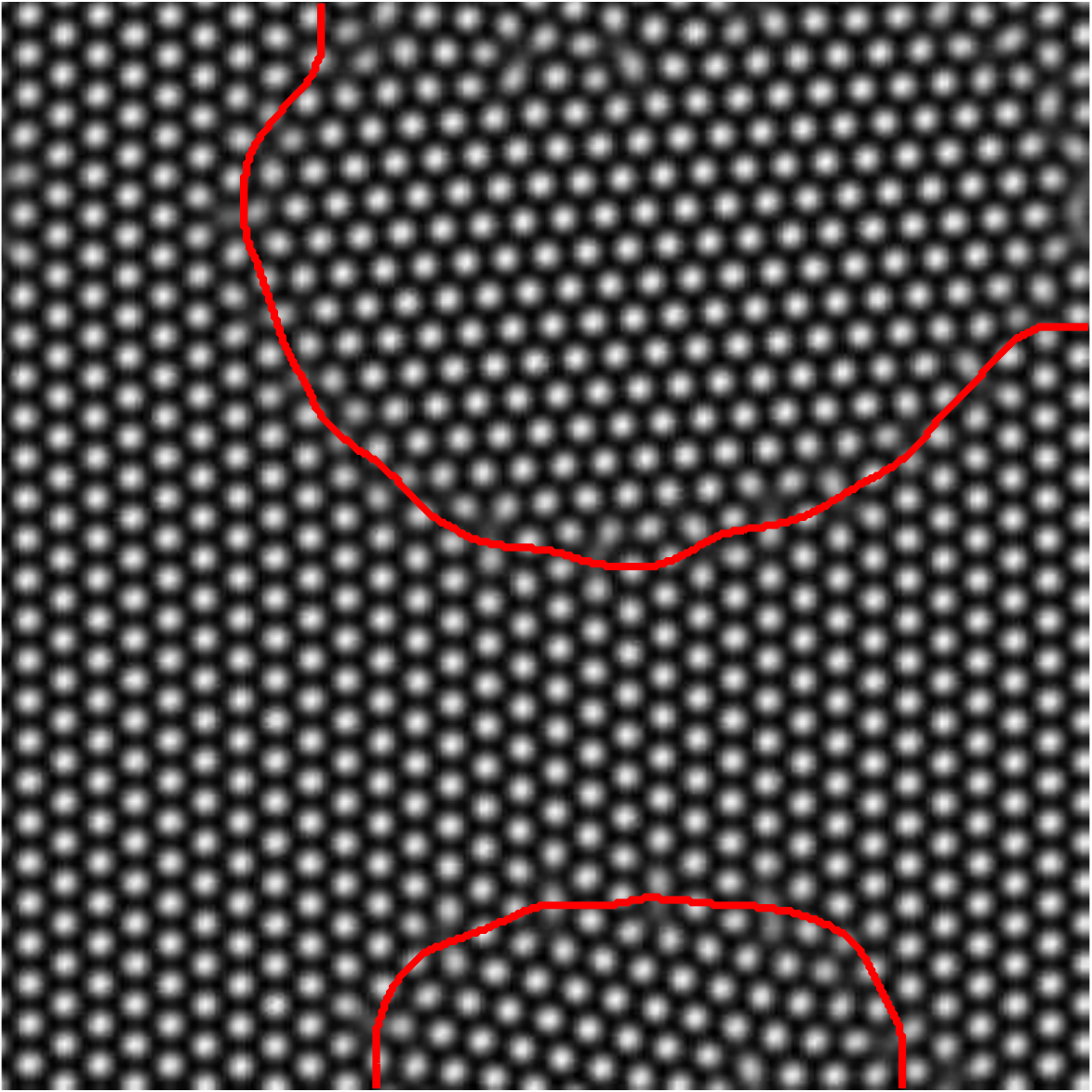}
	\includegraphics[width=0.215\textwidth]{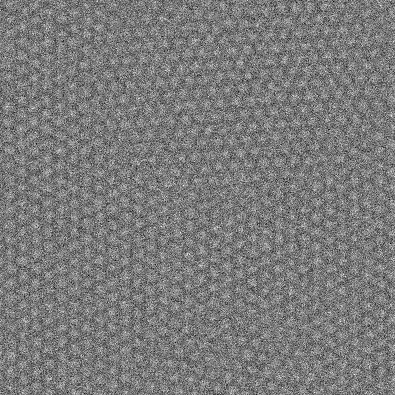}
	\includegraphics[width=0.215\textwidth]{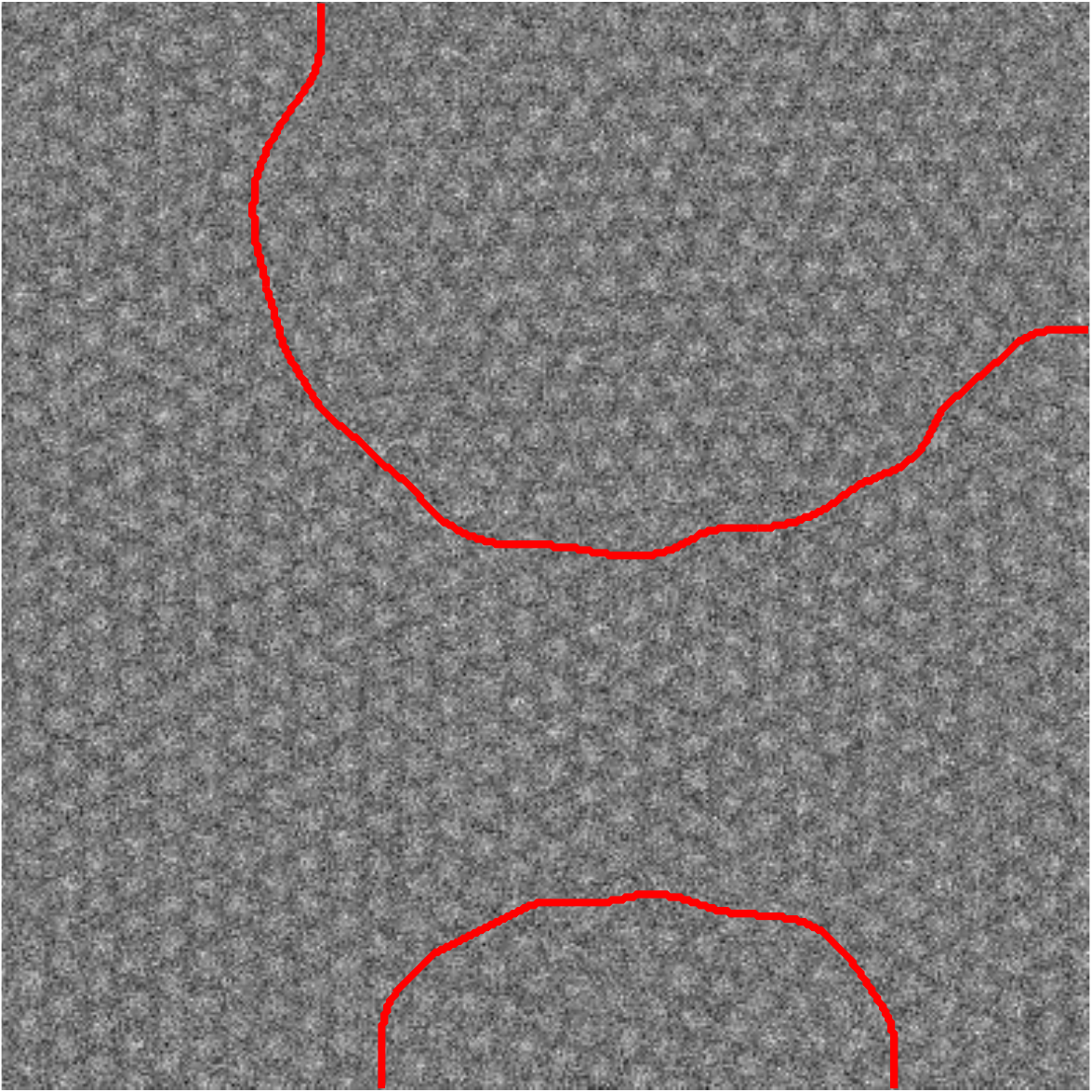}
 	\\ \vspace{0.1em} 
 	\includegraphics[width=0.215\textwidth]{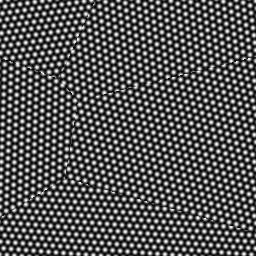}
	\includegraphics[width=0.215\textwidth]{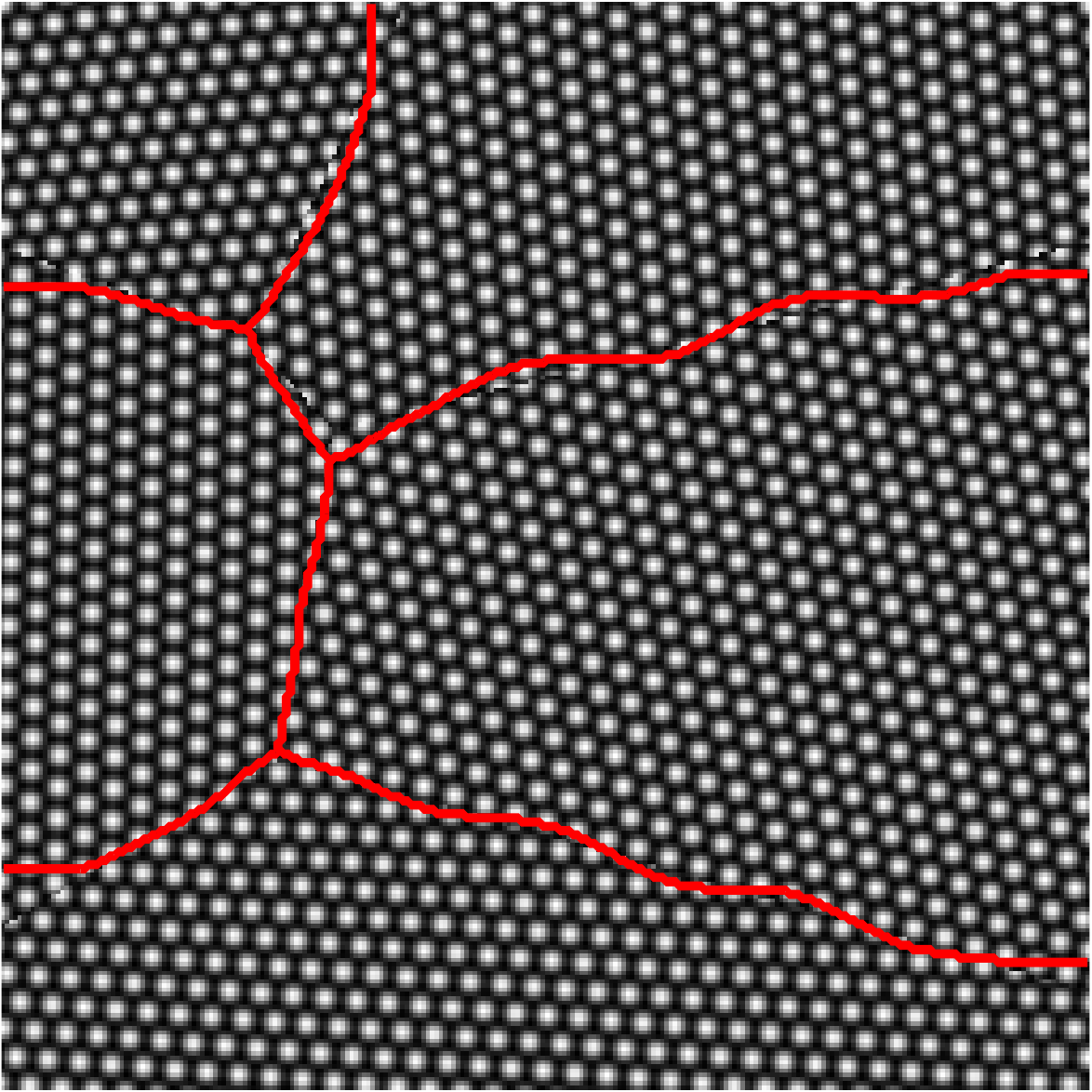}
	\includegraphics[width=0.215\textwidth]{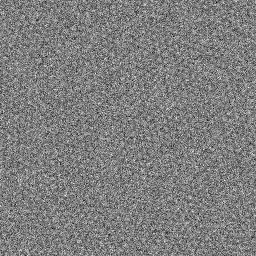}
	\includegraphics[width=0.215\textwidth]{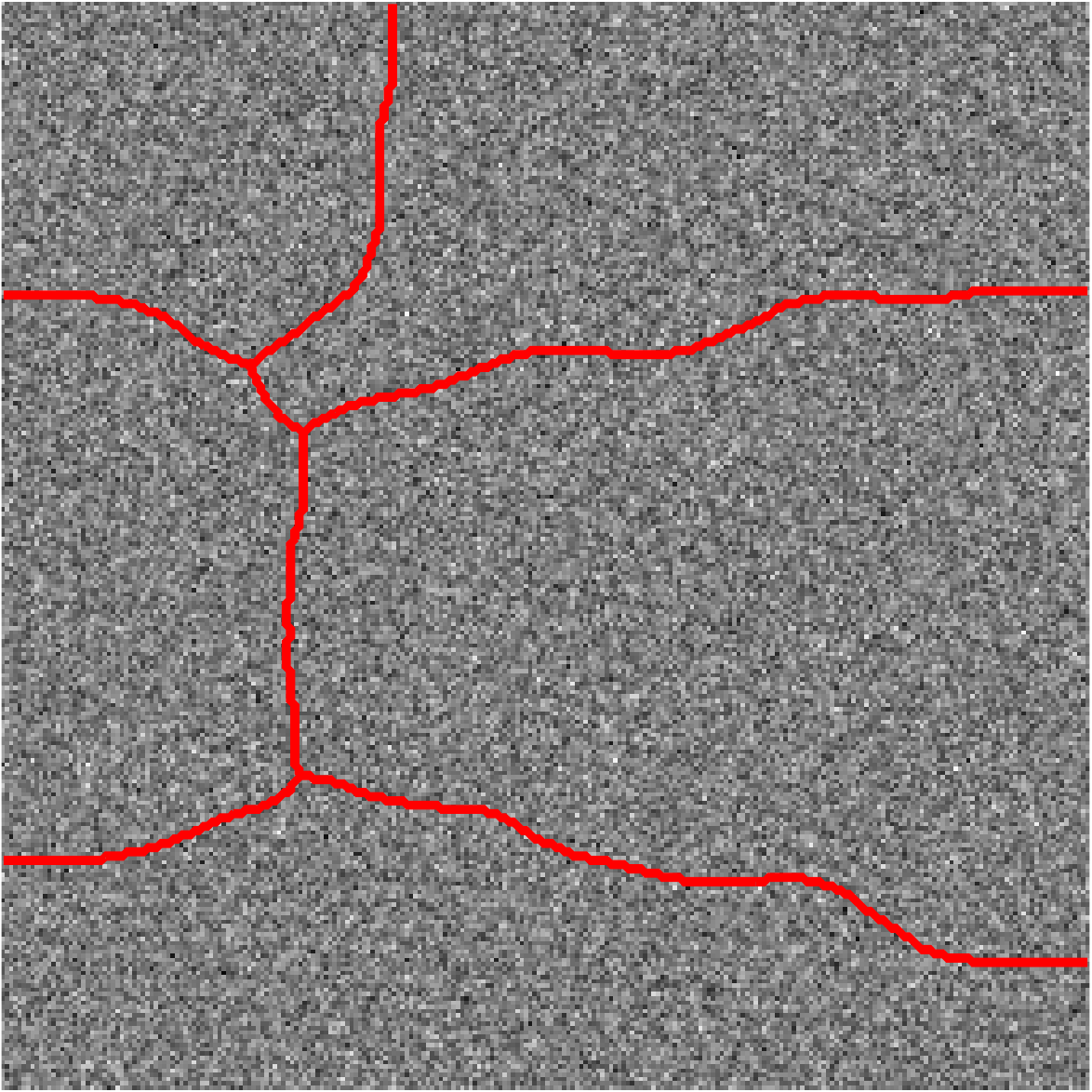}
	\\ \vspace{0.1em}
	\includegraphics[width=0.215\textwidth]{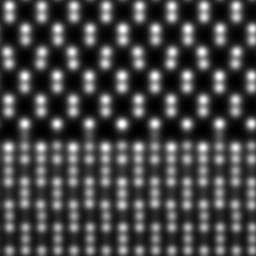}
	\includegraphics[width=0.215\textwidth]{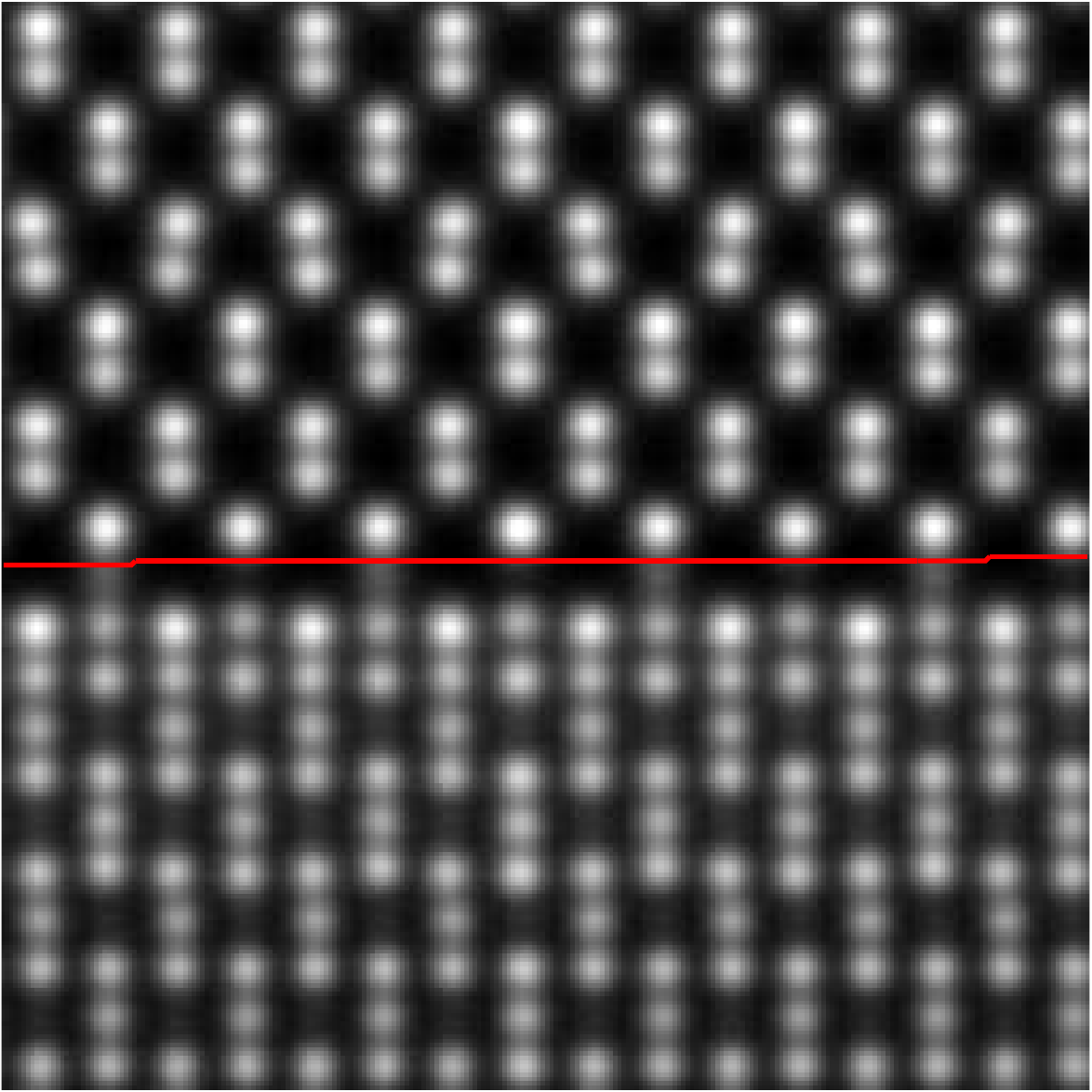}
	\includegraphics[width=0.215\textwidth]{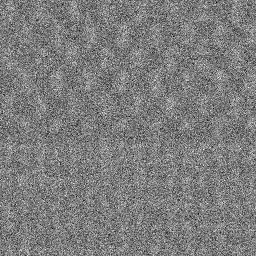}
	\includegraphics[width=0.215\textwidth]{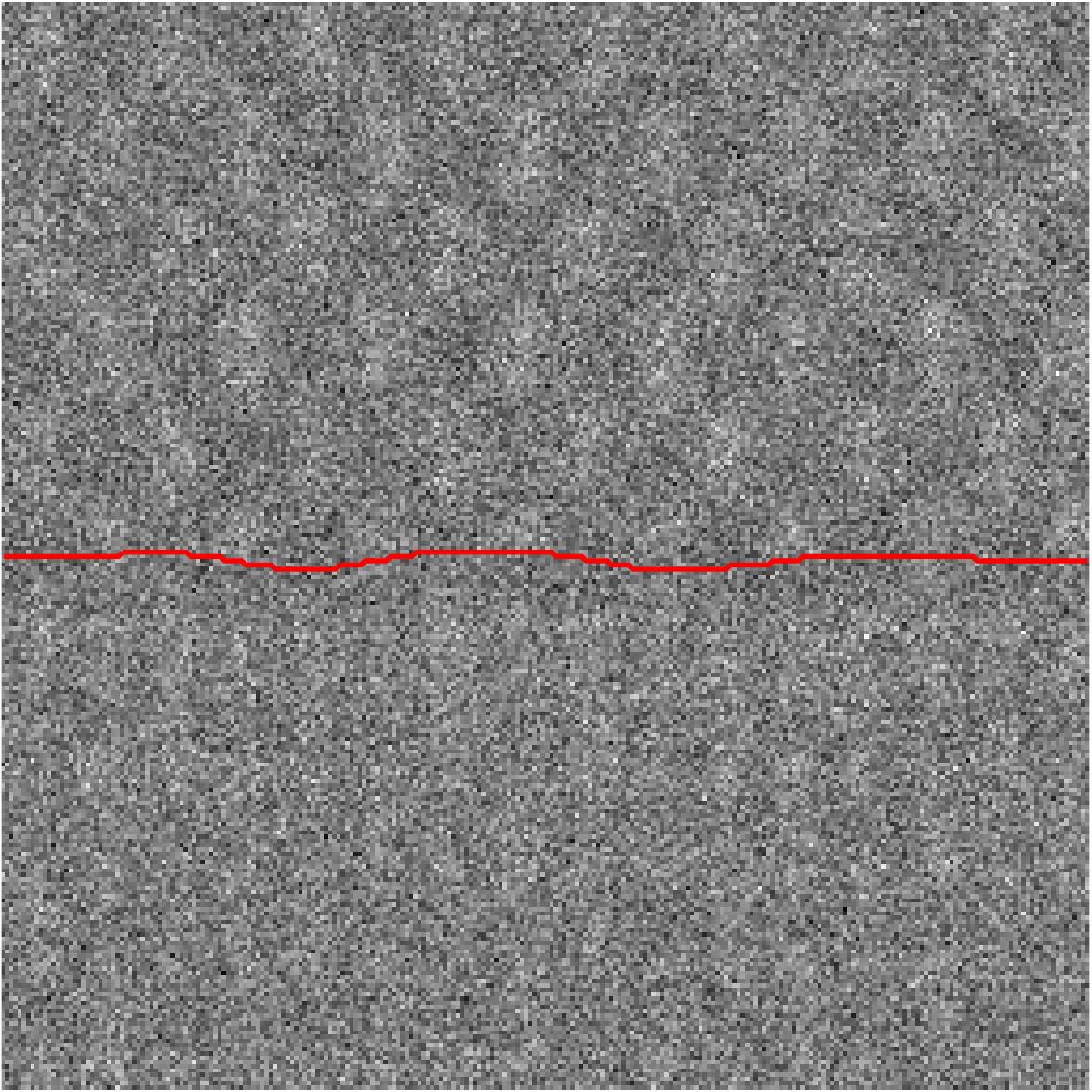}
\end{center}
   \caption{Segmentation of crystals without (left) and with (right) noise, computed by the proposed method using the 2D-FFT modulus \eqref{eq:fft} and visualized as boundary curves (red). Bottom row images courtesy of \textit{Paul Voyles}.}
\label{fig:crystals}
\end{figure*}

Figure~\ref{fig:crystals} shows segmentation results obtained by the proposed method with a 2D-FFT modulus based feature extractor of sizes $s=15$ (rows 1-3) and $s=20$ (last row). The crystals in the first three rows consist of regions differing only in crystal orientation. From visual inspection, the results for the noise-free images are exact up to inter-atomic distance. In the first row, despite the noisy grain (third column) begin hardly recognizable due to high noise power (Gaussian noise with a standard deviation of 100\% of the maximum noise-free image intensity), the segmentation deviates little from that of the noise-free grain. Similar results are observed for the three-phase scenario (second row). A lower noise power (66\%) was chosen, because otherwise the small bottom grain could not be detected, likely due to its small size compared to the window size. The multi-phase segmentation also works very well for five regions (third row) under the presence of very strong noise (100\%). Furthermore, as seen in the bottom row of Figure~\ref{fig:crystals}, the proposed Fourier-based segmentation is feasible and robust to large amounts of noise (100\%), even if the individual grains have entirely different crystal lattices. This is a type of material of practical relevance to material scientists that cannot be handled by the stencil based methods \cite{BeRaRu07,BoBeRu10,ElWi14a}.

%-------------------------------------------------------------------------
\section{Conclusions}
We have discussed a variational framework for multi-phase image segmentation based on structural information from high-dimensional local features. The framework imposes no special constraints on the used indicator functions, except that they are suitable for structure discrimination in the sense that they should be roughly homogeneous inside the structures of interest and provide some contrast across the different regions of interest. A robust initialization strategy for the segmentation algorithm was presented in this context, based on dimension reduction and decorrelation via PCA, as well as edgeness detection and clustering. Numerical results for two applications were presented. For texture segmentation, the proposed framework provides very competitive results, including state-of-the-art performance on the Prague benchmark. Using the 2D-FFT as feature extractor, robust and unsupervised crystal segmentation can be achieved, including segmentation of crystals with entirely different structure from extremely noisy data and without a-priori information about the crystals. We would like to point out that the proposed method can also be applied directly to high-dimensional data, for instance in spectroscopy.

The source code of the proposed method, including executables reproducing all presented results, is available at \url{http://nmevenkamp.github.io/pcams}.

\paragraph{Acknowledgments}
The authors thank Paul Voyles for providing simulated STEM images of a two-phase crystal.

{\small
\bibliographystyle{ieee}
\bibliography{bibtex_cameraReady}

\begin{thebibliography}{10}\itemsep=-1pt

\bibitem{AmFuPa00}
L.~Ambrosio, N.~Fusco, and D.~Pallara.
\newblock {\em Functions of bounded variation and free discontinuity problems}.
\newblock Oxford Mathematical Monographs. Oxford University Press, New York,
  2000.

\bibitem{ArMaFo11}
P.~Arbelaez, M.~Maire, C.~Fowlkes, and J.~Malik.
\newblock Contour detection and hierarchical image segmentation.
\newblock {\em PAMI}, 33(5):898--916, 2011.

\bibitem{BaGi96}
J.~Barba and J.~Gil.
\newblock An iterative algorithm for cell segmentation using short-time
  {F}ourier transform.
\newblock {\em J. Microsc.}, 184(2):127--132, 1996.

\bibitem{BeRaRu07}
B.~Berkels, A.~R{\"{a}}tz, M.~Rumpf, and A.~Voigt.
\newblock Extracting grain boundaries and macroscopic deformations from images
  on atomic scale.
\newblock {\em J. Sci. Comput.}, 35(1):1--23, 2008.

\bibitem{BoBeRu10}
M.~Boerdgen, B.~Berkels, M.~Rumpf, and D.~Cremers.
\newblock Convex relaxation for grain segmentation at atomic scale.
\newblock In {\em VMV}, 2010.

\bibitem{ChPo11}
A.~Chambolle and T.~Pock.
\newblock A first-order primal-dual algorithm for convex problems with
  applications to imaging.
\newblock {\em J. Math. Imaging Vision}, 40(1):120--145, 2011.

\bibitem{ChKa02}
D.~Charalampidis and T.~Kasparis.
\newblock Wavelet-based rotational invariant roughness features for texture
  classification and segmentation.
\newblock {\em TIP}, 11(8):825--837, 2002.

\bibitem{ChJiSu01}
H.-D. Cheng, X.~Jiang, Y.~Sun, and J.~Wang.
\newblock Color image segmentation: advances and prospects.
\newblock {\em Pattern recognition}, 34(12):2259--2281, 2001.

\bibitem{CoXuGr12}
M.~D. Collins, J.~Xu, L.~Grady, and V.~Singh.
\newblock Random walks based multi-image segmentation: Quasiconvexity results
  and {GPU}-based solutions.
\newblock In {\em CVPR}, 2012.

\bibitem{DiHe04}
C.~Ding and X.~He.
\newblock K-means clustering via principal component analysis.
\newblock In {\em ICML}, 2004.

\bibitem{DrStMi09}
S.~Drabycz, R.~G. Stockwell, and J.~R. Mitchell.
\newblock Image texture characterization using the discrete orthonormal
  {S}-transform.
\newblock {\em J. Digit. Imaging}, 22(6):696--708, 2009.

\bibitem{ElWi14a}
M.~Elsey and B.~Wirth.
\newblock Fast automated detection of crystal distortion and crystal defects in
  polycrystal images.
\newblock {\em Multiscale Modeling \& Simulation}, 12(1):1--24, 2014.

\bibitem{Felzenszwalb2004}
P.~F. Felzenszwalb and D.~P. Huttenlocher.
\newblock Efficient graph-based image segmentation.
\newblock {\em IJCV}, 59(2):167--181, 2004.

\bibitem{FrReHo12}
M.~M. Fraz, P.~Remagnino, A.~Hoppe, B.~Uyyanonvara, A.~R. Rudnicka, C.~G. Owen,
  and S.~A. Barman.
\newblock Blood vessel segmentation methodologies in retinal images--a survey.
\newblock {\em Comput. Methods Programs Biomed.}, 108(1):407--433, 2012.

\bibitem{HaMi14}
M.~Haindl and S.~Mike\u{s}.
\newblock Unsupervised image segmentation contest.
\newblock In {\em ICPR}, 2014.

\bibitem{Haindl2009}
M.~Haindl, S.~Mike\u{s}, and P.~Pudil.
\newblock Unsupervised hierarchical weighted multi-segmenter.
\newblock In J.~A. Benediktsson, J.~Kittler, and F.~Roli, editors, {\em
  Multiple Classifier Systems}, volume 5519 of {\em Lecture Notes in Computer
  Science}, pages 272--282. Springer Berlin Heidelberg, 2009.

\bibitem{HaFeBa13}
Y.~Han, X.-C. Feng, and G.~Baciu.
\newblock Variational and {PCA} based natural image segmentation.
\newblock {\em Pattern Recognition}, 46(7):1971--1984, 2013.

\bibitem{IlWh11}
D.~E. Ilea and P.~F. Whelan.
\newblock Image segmentation based on the integration of colour--texture
  descriptors---a review.
\newblock {\em Pattern Recognition}, 44(10):2479--2501, 2011.

\bibitem{KaLe97}
N.~Kambhatla and T.~K. Leen.
\newblock Dimension reduction by local principal component analysis.
\newblock {\em Neural Comput.}, 9(7):1493--1516, 1997.

\bibitem{KrKrZi09}
H.-P. Kriegel, P.~Kr\"{o}ger, and A.~Zimek.
\newblock Clustering high-dimensional data: A survey on subspace clustering,
  pattern-based clustering, and correlation clustering.
\newblock {\em ACM Trans. Knowl. Discov. Data}, 3(1):1:1--1:58, Mar. 2009.

\bibitem{LiNgZe10}
F.~Li, M.~K. Ng, T.~Y. Zeng, and C.~Shen.
\newblock A multiphase image segmentation method based on fuzzy region
  competition.
\newblock {\em SIAM J. Imaging Sci.}, 3(3):277--299, 2010.

\bibitem{LiBiPl12}
J.~Li, J.~M. Bioucas-Dias, and A.~Plaza.
\newblock Spectral--spatial hyperspectral image segmentation using subspace
  multinomial logistic regression and {M}arkov random fields.
\newblock {\em TGRS}, 50(3):809--823, 2012.

\bibitem{LiWa06}
X.~Liu and D.~Wang.
\newblock Image and texture segmentation using local spectral histograms.
\newblock {\em TIP}, 15(10):3066--3077, Oct 2006.

\bibitem{Mi86}
C.~Michelot.
\newblock A finite algorithm for finding the projection of a point onto the
  canonical simplex of $\mathbb{R}^n$.
\newblock {\em J. Optim. Theory Appl.}, 50(1):195--200, 1986.

\bibitem{MuSh89}
D.~Mumford and J.~Shah.
\newblock Optimal approximations by piecewise smooth functions and associated
  variational problems.
\newblock {\em Commun. Pure Appl. Math.}, 42(5):577--685, 1989.

\bibitem{PaFe03}
B.~Parker and D.~D. Feng.
\newblock Variational segmentation and {PCA} applied to dynamic {PET} analysis.
\newblock In {\em Pan-Sydney Area Workshop on Visual Information Processing},
  2003.

\bibitem{PaHaLi04}
L.~Parsons, E.~Haque, and H.~Liu.
\newblock Subspace clustering for high dimensional data: A review.
\newblock {\em SIGKDD Explor. Newsl.}, 6(1):90--105, 2004.

\bibitem{ReDu93}
T.~R. Reed and J.~H. DuBuf.
\newblock A review of recent texture segmentation and feature extraction
  techniques.
\newblock {\em CVGIP: Image understanding}, 57(3):359--372, 1993.

\bibitem{TaZhSh13}
C.~Tai, X.~Zhang, and Z.~Shen.
\newblock Wavelet frame based multiphase image segmentation.
\newblock {\em SIAM J. Imaging Sci.}, 6(4):2521--2546, 2013.

\bibitem{MaPoHe09}
L.~J. van~der Maaten, E.~O. Postma, and H.~J. van~den Herik.
\newblock Dimensionality reduction: A comparative review.
\newblock {\em JMLR}, 10(1-41):66--71, 2009.

\bibitem{VaSa12}
S.~R. Vantaram and E.~Saber.
\newblock Survey of contemporary trends in color image segmentation.
\newblock {\em J. Electron. Imaging}, 21(4):040901--1--040901--28, 2012.

\bibitem{WeHiDu96}
T.~P. Weldon, W.~E. Higgins, and D.~F. Dunn.
\newblock Efficient {G}abor filter design for texture segmentation.
\newblock {\em Pattern Recognition}, 29(12):2005--2015, 1996.

\bibitem{YuWaChe15}
J.~Yuan, D.~Wang, and A.~Cheriyadat.
\newblock Factorization-based texture segmentation.
\newblock {\em TIP}, 24(11):3488--3497, November 2015.

\bibitem{ChristopherZach08}
C.~Zach, D.~Gallup, J.-M. Frahm, and M.~Niethammer.
\newblock Fast global labeling for real-time stereo using multiple plane
  sweeps.
\newblock In {\em VMV}, 2008.

\end{thebibliography}
}

\end{document}